# New Results for the MAP Problem in Bayesian Networks

Cassio P. de Campos

*Istituto Dalle Molle di Studi sull'Intelligenza Artificiale (IDSIA)*
*Galleria 2, Manno 6928, Switzerland*
*cassio@idsia.ch*

**Abstract**

This paper presents new results for the (partial) maximum a posteriori (MAP) problem in Bayesian networks, which is the problem of querying the most probable state configuration of some of the network variables given evidence. First, it is demonstrated that the problem remains hard even in networks with very simple topology, such as binary polytrees and simple trees (including the Naive Bayes structure). Such proofs extend previous complexity results for the problem. Inapproximability results are also derived in the case of trees if the number of states per variable is not bounded. Although the problem is shown to be hard and inapproximable even in very simple scenarios, a new exact algorithm is described that is empirically fast in networks of bounded treewidth and bounded number of states per variable. The same algorithm is used as basis of a *Fully Polynomial Time Approximation Scheme* for MAP under such assumptions. Approximation schemes were generally thought to be impossible for this problem, but we show otherwise for classes of networks that are important in practice. The algorithms are extensively tested using some well-known networks as well as random generated cases to show their effectiveness.

## 1. Introduction

A Bayesian network (BN) is a probabilistic graphical model that relies on a structured dependency among random variables to represent a joint probability distribution in a compact and efficient manner. It is composed of a directed acyclic graph (DAG) where nodes are associated to random variables and conditional probability distributions are defined for variables given their parents in the graph. One of the hardest inference problems in BNs is the *maximum a posteriori* (or MAP) problem, where one looks for states of some variables that maximize their joint probability, given some other variables as evidence (there may exist variables that are neither queried nor part of the evidence). This problem is known to be $NP^{PP}$-complete in the general case and NP-complete for polytrees [1, 2]. Thus, algorithms usually take large amount of time to solve MAP even in small networks. Approximating MAP in polytrees is also NP-hard. However, such hardness results are derived for networks with a large number of states per variable, which is not the most common situation in many practical problems. In this paper we consider the case where the number of states per variable is bounded. We prove that the problem remains hard even in binary polytrees and simple trees, using reductions from both the *satisfiability* and the *partition* problems, but we also show that there is a *Fully Polynomial Time Approximation Scheme* (FPTAS) whenever the treewidth and number of states are bounded, so we may expect fast algorithms for MAP with a small approximation error under such assumptions. A new exact algorithm is presented, which also delivers approximations with theoretically bounded errors. Empirical results show that this algorithm surpasses a state-of-the-art method for the same problem. Fast algorithms for MAP imply in fast algorithms for other related problems, for example inferences in decision networks and influence diagrams, besides the great interest in the MAP problem itself. Hence, this paper makes important steps in these directions.

## 2. Background

In this section we formally define the networks, the problem, and the algorithmic techniques that are used to prove the new complexity results as well as to devise the new algorithm for MAP. We assume that the reader is familiar with basic notions of complexity theory and approximation algorithms (for more details, see for example [3, 4, 5]) and basic concepts of Bayesian networks [6, 7, 8].

**Definition 1** *A Bayesian network (BN) $\mathcal{N}$ is defined by a triple $(\mathcal{G}, \mathcal{X}, \mathcal{P})$, where $\mathcal{G} = (\mathbf{V}_\mathcal{G}, \mathbf{E}_\mathcal{G})$ is a directed acyclic graph with nodes $\mathbf{V}_\mathcal{G}$ associated (in a one-to-one mapping) to random variables $\mathcal{X} = \{X_1, \ldots, X_n\}$ over discrete domains $\{\Omega_{X_1}, \ldots, \Omega_{X_n}\}$ and $\mathcal{P}$ is a collection of probability values $p(x_i|\pi_{X_i}) \in \mathcal{Q}$,[1] with $\sum_{x_i \in \Omega_{X_i}} p(x_i|\pi_{X_i}) = 1$, where $x_i \in \Omega_{X_i}$ is a category or state of $X_i$ and $\pi_{X_i} \in \times_{X \in \Pi_{X_i}} \Omega_X$ a complete instantiation for the parents $\Pi_{X_i}$ of $X_i$ in $\mathcal{G}$. Furthermore, every variable is conditionally independent of its non-descendants given its parents.*

Given its independence assumptions among variables, the joint probability distribution represented by a BN $(\mathcal{G}, \mathcal{X}, \mathcal{P})$ is obtained by $p(\mathbf{x}) = \prod_i p(x_i|\pi_{X_i})$, where $\mathbf{x} \in \Omega_\mathcal{X}$ and all states $x_i, \pi_{X_i}$ (for every $i$) agree with $\mathbf{x}$.

For ease of expose, we denote the singletons $\{X_i\}$ and $\{x_i\}$ respectively as $X_i$ and $x_i$. Nodes of the graph and their associated random variables are used interchanged. Uppercase letters are used for random variables and lowercase letters for their corresponding states. Bold letters are employed for vectors/sets. We denote by $z(\mathbf{X})$ the product of the cardinality of the variables $\mathbf{X} \subseteq \mathcal{X}$, that is, $z(\mathbf{X}) = \prod_{X_i \in \mathbf{X}} z(X_i)$, where $z(X_i) = |\Omega_{X_i}|$ is the number of states (cardinality) of $X_i$. We assume $z(\emptyset) = 1$. We use simply $z_i$ to denote $z(X_i)$. The input size of a BN is given by the sum of the sizes to specify the local conditional probability distributions and the space to describe the graph, that is, $\text{size}(\mathcal{N}) \in \Theta(|\mathbf{E}_\mathcal{G}|) + \sum_i (z_i - 1) \prod_{X_j \in \Pi_{X_i}} z_j \in \sum_i \Theta(z(X_i \cup \Pi_{X_i}))$,[2][3] as $\Theta(|\mathbf{E}_\mathcal{G}|)$ is clearly dominated by the summation $\sum_i \Theta(z(X_i \cup \Pi_{X_i}))$. Note that $\text{size}(\mathcal{N}) \in \Omega(n)$ and $\text{size}(\mathcal{N}) \in \Omega(z_{max})$, where $z_{max} = \max_{X_i \in \mathcal{X}} z_i$.

The belief updating (BU) problem concerns the computation of $p(\mathbf{x}|\mathbf{e})$, for $\mathbf{x} \in \Omega_\mathbf{X}$ and $\mathbf{e} \in \Omega_\mathbf{E}$, with $\mathbf{X} \cup \mathbf{E} \subseteq \mathcal{X}$ and $\mathbf{X} \cap \mathbf{E} = \emptyset$. It is known that the decision version of this problem (we denote it by Decision-BU), which can be stated as *"is it true that $p(\mathbf{x}|\mathbf{e}) > r$, for a given rational $r$"*, is PP-hard [9], using a reduction from MAJ-SAT (majority satisfiability). As $p(\mathbf{x}|\mathbf{e}) = \frac{p(\mathbf{x}, \mathbf{e})}{p(\mathbf{e})}$, in the following we discuss how to compute the query $p(\mathbf{x}')$, and the terms $p(\mathbf{x}, \mathbf{e})$ and $p(\mathbf{e})$ are obtained analogously by letting $\mathbf{x}' = \mathbf{x}, \mathbf{e}$[4] and $\mathbf{x}' = \mathbf{e}$, respectively:

$$p(\mathbf{x}') = \sum_{\mathbf{y} \in \Omega_{\mathcal{X} \setminus \mathbf{x}'}} p(\mathbf{y}, \mathbf{x}') = \sum_{\mathbf{y} \in \Omega_{\mathcal{X} \setminus \mathbf{x}'}} \prod_{X_i \in \mathcal{X}} p(x_i|\pi_{X_i}), \tag{1}$$

where $x_i \in \Omega_{X_i}$ and $\pi_{X_i} = \times_{X \in \Pi_{X_i}} \Omega_X$ (for all $i$) agree with $\mathbf{y}, \mathbf{x}'$. Eq. (1) is a summation with exponentially many terms, each one requiring less than $n$ multiplications. However, we can compute this huge summation in some specific ordering to save time. Some definitions are required here. The moral graph of a network $\mathcal{N} = (\mathcal{G}, \mathcal{X}, \mathcal{P})$ (denoted $\mathcal{G}_m$) is obtained from $\mathcal{G}$ by connecting the nodes of $\mathcal{G}$ that have a common child (marrying parents), and then dropping the direction of all the arcs. Well-known inference methods [10, 11] use a tree decomposition of $\mathcal{G}_m$ to propagate results and speed up computations.

**Definition 2** *Given a graph $\mathcal{G} = (\mathbf{X}_\mathcal{G}, \mathbf{E}_\mathcal{G})$, a tree decomposition $\mathcal{T}$ of $\mathcal{G}$ is a pair $(\mathbf{C}, T)$, where $\mathbf{C} = \{\mathbf{C}_1, \ldots \mathbf{C}_N\}$ is a family of non-empty subsets of $\mathbf{X}_\mathcal{G}$, and $T$ is a tree where the nodes are associated (in a one-to-one mapping) to the subsets $\mathbf{C}_i$, satisfying the following properties: (i) $\bigcup_i \mathbf{C}_i = \mathbf{X}_\mathcal{G}$; (ii) For every edge $E \in \mathbf{E}_\mathcal{G}$, there is a subset $\mathbf{C}_i$ that contains both extremes of $E$; (iii) If both $\mathbf{C}_i$ and $\mathbf{C}_j$ (with $i \neq j$) contain a vertex $X$, then all nodes of the tree in the (unique) path between $\mathbf{C}_i$ and $\mathbf{C}_j$ contain $X$ as well.*

Let $\mathcal{T} = (\mathbf{C}, T)$ be a tree decomposition of $\mathcal{G}_m$, with $\mathbf{C} = \{\mathbf{C}_1, \ldots, \mathbf{C}_{n'}\}$ and $n' < n$ (this does not imply any loss of generality [12]). Elect a node and assume all edges of $T$ point towards the opposite direction of it. Without loss of generality, let $\mathbf{C}_1$ be this node and $\mathbf{C}_1, \ldots, \mathbf{C}_{n'}$ be a topological order with respect to this graph, that is, the path between $\mathbf{C}_1$ and $\mathbf{C}_j$ in $T$ does not contain any $\mathbf{C}_{j'}$ with $j' > j$. Let $\mathbf{C}_{j_p} = \Pi_{\mathbf{C}_j}$ be the only parent of $\mathbf{C}_j$ in the tree and $\Lambda_{\mathbf{C}_j}$ be the children of $\mathbf{C}_j$. We say that $\mathcal{T}' = (\mathbf{C}, T)$ is a *binary* decomposition if $|\Lambda_{\mathbf{C}_j}| \leq 2$, for every $\mathbf{C}_j$ [13]. Note that it is easy to obtain a binary decomposition $\mathcal{T}'$ from $\mathcal{T}$: include additional nodes $\mathbf{C}_{j,k}$ for each $\mathbf{C}_j$ that has more than two children (in number equal to $|\Lambda_{\mathbf{C}_j}| - 1$) such that: (i) the variables inside each $\mathbf{C}_{j,k}$ are the same variables as those inside $\mathbf{C}_j$; (ii) the nodes $\mathbf{C}_{j,k}$ form a chain, where the root of the chain is $\mathbf{C}_{j,1} = \mathbf{C}_j$ and each $\mathbf{C}_{j,k}$ ($k \geq 1$) has $\mathbf{C}_{j,k+1}$ and one of the original children of $\mathbf{C}_j$ as its children. This transformation preserves the tree decomposition properties and reduces the number of children of each $\mathbf{C}_{j,k}$ to exactly two. Moreover, the maximum number of variables inside a single $\mathbf{C}_j$ is not changed (as we have just replicated $\mathbf{C}_j$ into the elements $\mathbf{C}_{j,k}$), and the total number of nodes in the new tree is less than $2n$. Binary tree decompositions are useful later during some derivations.

---

[1] $\mathcal{Q}$ denotes the rational numbers, which are supposed to be given by two integers defining the numerator and the denominator of the corresponding fractions.

[2] If probability values repeat in a systematic way, one could represent the network in a more compact form. We do not consider such situation.

[3] We employ Knuth's asymptotic notation $\Omega(f)$, $O(f)$ and $\Theta(f)$. The reader shall not confuse the state spaces denoted by $\Omega_X$ (with a subscript) and the asymptotic notation $\Omega(f)$. We use the notation $g \in \Omega(f)$ when $g$ is $\Omega(f)$, that is, $f$ is an asymptotic lower bound for $g$, and so on.

[4] The comma shall be seen as a concatenation operator, that is, $p(\mathbf{x}') = p(\mathbf{x}, \mathbf{e}) = p(\mathbf{x} \wedge \mathbf{e})$ is the probability of observing altogether the elements in $\mathbf{x}'$.

Let $\mathbf{X}_j^{\text{last}} = \mathbf{C}_j \setminus \mathbf{C}_{j_p}$ be the set of nodes of $\mathcal{G}$ in $\mathbf{C}_j$ that do not appear in $\mathbf{C}_{j_p}$ (they also do not appear in any other node towards $\mathbf{C}_1$ because of the tree decomposition properties). We have the following recursion, which is processed in reverse topological order (from $j = n'$ to 1):

$$p(\mathbf{u}_{\mathbf{C}_j} | \mathbf{v}_{\mathbf{C}_j}) = \sum_{\Omega_{\mathbf{X}_j^{\text{last}} \setminus \mathbf{X}'}} \prod_{X_i \in \mathbf{X}_j^{\text{proc}}} p(x_i | \pi_{X_i}) \prod_{\mathbf{C}_{j'} \in \Lambda_{\mathbf{C}_j}} p(\mathbf{u}_{\mathbf{C}_{j'}} | \mathbf{v}_{\mathbf{C}_{j'}}), \qquad (2)$$

where $\mathbf{X}_j^{\text{proc}} = \{X_i \in \mathbf{C}_j : (X_i \cup \Pi_{X_i}) \cap \mathbf{X}_j^{\text{last}} \neq \emptyset\}$ is the set of variables whose local probability functions were not processed yet (but need to be) in order to sum out over the elements $\mathbf{X}_j^{\text{last}} \setminus \mathbf{X}'$. Furthermore, $\mathbf{U}_{\mathbf{C}_j} = (\mathbf{X}_j^{\text{proc}} \cup \bigcup_{\mathbf{C}_{j'} \in \Lambda_{\mathbf{C}_j}} \mathbf{U}_{\mathbf{C}_{j'}}) \setminus \mathbf{X}_j^{\text{last}}$ is composed of elements of $\mathbf{C}_j$ and descendants that are also present in the parent $\mathbf{C}_{j_p}$ and whose local probability distributions were already taken into account (they do appear in the left side of the conditioning bar in $\mathbf{C}_j$ or in its descendants), and finally $\mathbf{V}_{\mathbf{C}_j} = \mathbf{C}_j \setminus (\mathbf{U}_{\mathbf{C}_j} \cup \mathbf{X}_j^{\text{last}})$ are the variables that already appeared in the right side of the conditioning bar (but not in the left side nor they were summed out).

The recursion of Eq. (2) formalizes the engine behind well known algorithms (for instance, bucket elimination) for BU in BNs. Putting in words, the values $p(\mathbf{u}_{\mathbf{C}_{j'}} | \mathbf{v}_{\mathbf{C}_{j'}})$ represent the information that comes from the children of $\mathbf{C}_j$. They can be seen as functions over the domains $\Omega_{\mathbf{U}_{\mathbf{C}_{j'}} \cup \mathbf{V}_{\mathbf{C}_{j'}}}$ that come from independent subtrees and are multiplied altogether and by the probability functions that appear for the first time in $\mathbf{C}_j$ (if any). Then such functions are summed out over the variables that do not appear in the parent of the current $\mathbf{C}_j$ to build the information that will be "propagated" to the parent $\mathbf{C}_{j_p}$, that is, they are all used to construct the function over $\Omega_{\mathbf{U}_{\mathbf{C}_j} \cup \mathbf{V}_{\mathbf{C}_j}}$ defined by the values $p(\mathbf{u}_{\mathbf{C}_j} | \mathbf{v}_{\mathbf{C}_j})$. The recursion is evaluated for each $j$, and finally $p(\mathbf{u}_{\mathbf{C}_1}) = p(\mathbf{x}')$ (in this case $\mathbf{V}_{\mathbf{C}_1}$ is certainly empty). Note that $\mathbf{U}_{\mathbf{C}_j} \cup \mathbf{V}_{\mathbf{C}_j} \subseteq (\mathbf{C}_j \cap \Pi_{\mathbf{C}_j})$ and $\mathbf{U}_{\mathbf{C}_j} \cap \mathbf{V}_{\mathbf{C}_j} = \emptyset$. Because $p(\mathbf{u}_{\mathbf{C}_j} | \mathbf{v}_{\mathbf{C}_j})$ is evaluated for each instantiation of $\mathbf{u}_{\mathbf{C}_j} \in \Omega_{\mathbf{U}_{\mathbf{C}_j}}$ and $\mathbf{v}_{\mathbf{C}_j} \in \Omega_{\mathbf{V}_{\mathbf{C}_j}}$ that agrees with $\mathbf{x}'$, there are $z((\mathbf{U}_{\mathbf{C}_j} \cup \mathbf{V}_{\mathbf{C}_j}) \setminus \mathbf{X}') \leq z((\mathbf{C}_j \cap \Pi_{\mathbf{C}_j}) \setminus \mathbf{X}')$ numbers to be computed. Each such computation requires at most $z(\mathbf{X}_j^{\text{last}})$ (the summation has at most this number of terms) times at most $|\Lambda_{\mathbf{C}_j}| + 1$ multiplications. Thus, the total running time is at most

$$\sum_j (1 + |\Lambda_{\mathbf{C}_j}|) \cdot z(\mathbf{U}_{\mathbf{C}_j} \cup \mathbf{V}_{\mathbf{C}_j}) \cdot z(\mathbf{X}_j^{\text{last}}) \leq \sum_j (1 + |\Lambda_{\mathbf{C}_j}|) \cdot z(\mathbf{C}_j) \in O(n \cdot z_{max}^w), \qquad (3)$$

where $w = \max_j |\mathbf{C}_j|$. We have used the fact that $\sum_j |\Lambda_{\mathbf{C}_j}| \leq n'$, and $n' \in O(n)$. This computational time is exponential in the size of the sets $\mathbf{C}_j$ of the tree, so it is reasonable to look for a decomposition with small *treewidth*:

**Definition 3** *The* treewidth *of a graph $\mathcal{G}$ w.r.t. the tree decomposition $\mathcal{T} = (\mathbf{C}, T)$ is the maximum number of nodes of $\mathcal{G}$ in a node of $T$ minus 1: $w(\mathcal{G}, \mathcal{T}) = -1 + \max_j |\mathbf{C}_j|$. Moreover, $w^*(\mathcal{G}) = \min_{\mathcal{T}} w(\mathcal{G}, \mathcal{T})$, with $\mathcal{T}$ ranging over all possible tree decompositions, is the* minimum treewidth *of a graph $\mathcal{G}$.*

Finding the tree decomposition with *minimum treewidth* is a NP-complete problem [14]. The best known approximation method achieves an $O(\log n)$ factor of the optimal [15]. In spite of that, some particular BNs deserve attention: $\mathcal{N} = (\mathcal{G}, \mathcal{X}, \mathcal{P})$ is called a *polytree* if the subjacent graph[5] of $\mathcal{G}$ has no cycles. A polytree $\mathcal{N}$ is further called a *tree* if each node of $\mathcal{G}$ has at most one parent. For trees and polytrees, Decision-BU is solvable in polynomial time [7]. As we see in Eq. (3), this is also true for any network of bounded treewidth. In fact the moral graph of a polytree may have a large treewidth if the number of parents of a variable is large. However, the input size would proportionally increase too, and the polynomial time on the input size would sustain. We do not discuss this case further for ease of expose.

---
[5] A subjacent graph of $\mathcal{G}$ is the graph obtained by dropping the direction of the arcs.

The MAP problem is to find an instantiation $\mathbf{x}' \in \Omega_\mathbf{X}$, with $\mathbf{X} \subseteq \mathcal{X} \setminus \mathbf{E}$, such that its probability is maximized, that is,

$$\mathbf{x}' = \underset{\mathbf{x} \in \Omega_\mathbf{X}}{\mathrm{argmax}}\, p(\mathbf{x}|\mathbf{e}) = \underset{\mathbf{x} \in \Omega_\mathbf{X}}{\mathrm{argmax}}\, \frac{p(\mathbf{x},\mathbf{e})}{p(\mathbf{e})} = \underset{\mathbf{x} \in \Omega_\mathbf{X}}{\mathrm{argmax}}\, p(\mathbf{x},\mathbf{e}), \qquad (4)$$

because $p(\mathbf{e})$ (assumed to be non-zero) is a constant with respect to the maximization. It is known that MAP queries are harder than BU queries (under the assumptions that P$\neq$NP and PP$\neq$NP$^{\mathrm{PP}}$). It is proved that the general MAP problem is NP$^{\mathrm{PP}}$-complete [1]. However, such proof assumes a general case of the problem, while many practical BNs have some structural properties that might alleviate the complexity. Two of them are very important with respect to the complexity of the problem: the cardinality of the variables involved in the network and the minimum treewidth of the moralized graph (which is expected, because this value also affects the BU complexity). The latter is considered in [1], and the problem is shown to be NP-complete and not approximable by a polynomial approximation scheme. We make use of a parametrized version of MAP to exploit these two characteristics.

**Definition 4** *Given a BN $\mathcal{N} = (\mathcal{G}, \mathcal{X}, \mathcal{P})$ where $z$ is the maximum cardinality of any variable and $w^*(\mathcal{G}_m) = w$ is the minimum treewidth of the moral graph of $\mathcal{G}$, $\mathbf{X} \subseteq \mathcal{X} \setminus \mathbf{E}$, a rational $r$ and an instantiation $\mathbf{e} \in \Omega_\mathbf{E}$, Decision-MAP-$z$-$w$ is the problem of deciding if there is $\mathbf{x} \in \Omega_\mathbf{X}$ such that $p(\mathbf{x}, \mathbf{e}) > r$. MAP-$z$-$w$ is the respective optimization version.[6] Furthermore, we denote by Decision-MAP-$\infty$-$w$ the MAP problem where $z$ can be asymptotically as large as the input size($\mathcal{N}$) (this is the same as having no bound, as we know that size($\mathcal{N}$) $\in \Omega(z)$).*

### 3. Complexity results

The results of this section show that MAP remains NP-complete even when restricted to very simple networks. Previously the hardness has been proved just for polytrees where the maximum cardinality of each variable was $\Omega(\mathrm{size}(\mathcal{N}))$ [1] (more specifically, the proof showed that Decision-MAP-$\infty$-$w$ is NP-hard for $w = 2$, because the cardinality could be as large as the number of clauses in the SAT problem used in the reduction, and the number of clauses of a SAT problem can be (asymptotically) as large as the corresponding input size. It was also shown that Decision-MAP-$z$-$\infty$ is NP-hard for $z = 2$ but $w$ unbounded). Here we show that the problem remains hard even when restricted to:

- Simple *binary* polytrees with at most two parents per node (which directly strengthens previous results).

- Trees with no bound on maximum cardinality but network topology as simple as a Naive Bayes structure [16] (in fact such version of the problem does not admit a polynomial approximation scheme, as we show that the optimization version is equivalent to the optimization version of MAXSAT [1]).

- Trees with bounded maximum cardinality and network topology as simple as a Hidden Markov Model structure [17] (this result shows that Decision-MAP is hard even in trees with bounded cardinality per variable).

Altogether these new complexity proofs strongly indicate that MAP problems are hard even when the underlying structure of the BN is very simple. First, Theorem 5 states the well-known fact that MAP is within NP when $w$ is fixed. Then Theorems 6, 8, and 10 present the hardness results.

**Theorem 5** *Decision-MAP-$z$-$w$ is in NP for any fixed $w$.*

**Proof** Pertinence in NP is trivial because Eq. (3) is polynomial in size($\mathcal{N}$) if the minimum treewidth is at most $w$ (there is a linear time algorithm to find a tree decomposition of minimum treewidth when $w$ is fixed [12]). So, given an instantiation $\mathbf{x}$, we can check whether $p(\mathbf{x}, \mathbf{e}) > r$ (or even $p(\mathbf{x}|\mathbf{e}) > r$) by Eq. (2) in polynomial time. $\square$

**Theorem 6** *Decision-MAP-$z$-$w$ is NP-hard even if $z = w = 2$.*

---

[6]In fact the results of this paper hold in both the unconditional (as presented in Def. 4) and the conditional formulation $p(\mathbf{x}|\mathbf{e}) > r$ for the Decision-MAP-$z$-$w$ problem, where the evidence in treated as conditioning information, because we deal with cases of bounded $w$ and computing $p(\mathbf{e})$ is a polynomial-time task.

**Proof** Hardness is shown using a reduction from *partition* problem, which is NP-hard [3] and can be stated as follows: *given a set of $m$ positive integers $s_1, \ldots, s_m$, is there a set $I \subset A = \{1, \ldots, m\}$ such that $\sum_{i \in I} s_i = \sum_{i \in A \setminus I} s_i$?* All the input is encoded using $b > 0$ bits.

Denote $S = \frac{1}{2} \sum_{i \in A} s_i$ and call *even partition* a subset $I \subset A$ that achieves $\sum_{i \in I} s_i = S$. To solve *partition*, we consider the rescaled problem (dividing every element by $S$), so as $v_i = \frac{s_i}{S} \leq 2$ are the elements and we look for a partition with sum equals to 1 (altogether the elements sum 2).

We construct (in polynomial time) a binary polytree (so $z_{max} = 2$) with $3m + 1$ nodes where the maximum number of parents of a node is 2, which implies that there is a tree decomposition of the moral graph with treewidth $w = 2$ (to see that, just take the same polytree and define the nodes $\mathbf{C}_j$ containing $X_j \cup \Pi_{X_j}$). The binary nodes are $\mathbf{X} = \{X_1, \ldots, X_m\}$, $\mathbf{Y} = \{Y_0, Y_1, \ldots, Y_m\}$ and $\mathbf{E} = \{E_1, \ldots, E_m\}$. We denote by $\{x_i^T, x_i^F\}$ the states of $X_i$ (similarly for $Y_i$ and $E_i$). The structure of the network is presented in Figure 1. Each $X_i \in \mathbf{X}$ has no parents and uniform distribution, each $E_i$ has $X_i$ as sole parent, with probability values defined as $p(e_i^T | x_i^F) = 1$ and $p(e_i^T | x_i^T) = t_i$ (the values for $e_i^F$ complement those to sum one), where $t_i$ is obtained by evaluating $2^{-v_i}$ up to $4b + 3$ bits of precision (and rounded up if necessary), that is, $t_i = 2^{-v_i} + \text{error}_i$, where $0 \leq \text{error}_i < 2^{-(4b+3)}$. Clearly $t_i$ can be computed in polynomial time and space in $b$ (this ensures that the specification of the Bayesian network, which requires rational numbers, is polynomial in $b$). Furthermore, note that $2^{-v_i} \leq t_i \leq 2^{-v_i} + \text{error}_i < 2^{-v_i + 2^{-4b}}$ (by Corollary 16, see appendix for details).

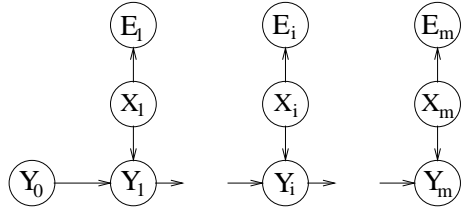

Figure 1: Network structure for the proof of Theorem 6.

$Y_0$ has no parents and $p(y_0^T) = 1$. For the nodes $Y_i \in \mathbf{Y}$, $1 \leq i \leq m$, the parents are $X_i$ and $Y_{i-1}$, and the probability values are $p(y_i^T | y_{i-1}^T, x_i^T) = t_i$, $p(y_i^T | y_{i-1}^T, x_i^F) = 1$, and $p(y_i^T | y_{i-1}^F, x_i) = 0$ for both states $x_i \in \Omega_{X_i}$.

Note that with this specification and given the Markov condition of the network, we have (for any given $\mathbf{x} \in \Omega_{\mathbf{X}}$) $p(y_m^T | \mathbf{x}) = p(\mathbf{e}^T | \mathbf{x}) = \prod_{i \in I} t_i$, where $I \subseteq A$ is the indicator of the elements such that $X_i$ is at the state $x_i^T$. Denote $\mathbf{t} = \prod_{i \in I} t_i$.

$$p(\mathbf{x}, \mathbf{e}^T, \neg y_m^T) = p(\neg y_m^T | \mathbf{x}) p(\mathbf{x}, \mathbf{e}^T) = p(\mathbf{x}) p(\mathbf{e}^T | \mathbf{x}) \left(1 - p(y_m^T | \mathbf{x})\right) = \frac{1}{2^m} \mathbf{t}(1 - \mathbf{t}). \tag{5}$$

This is a concave quadratic function on $\mathbf{t}$ with maximum at $2^{-1}$. Moreover, the value of $\mathbf{t}(1 - \mathbf{t})$ monotonically increases when $\mathbf{t}$ approaches one half (from both sides). For a moment, suppose that $t_i$ (defined in the previous paragraph) is exactly $2^{-v_i}$ (instead of an approximation of it as described). In this case,

$$\frac{1}{2^m} \mathbf{t}(1 - \mathbf{t}) = \frac{1}{2^m} 2^{-\sum_{i \in I} v_i} (1 - 2^{-\sum_{i \in I} v_i}),$$

which achieves the maximum of $\frac{1}{2^m} 2^{-1}(1 - 2^{-1})$ if and only if $\sum_{i \in I} v_i = 1$, which means that there is an even partition. This proof would be ended here if there was not the following consideration: we must show that the transformation is computed in polynomial time and the parameters of a BN are rational numbers, and computing $2^{-v_i}$ (needed to define the BN) might be an issue. For such purpose we employ an approximate version of $2^{-v_i}$ to define each $t_i$. The remainder of this proof addresses the question of how the numerical errors introduced in the definition of values $t_i$ interfere in the main result. Hence, note that if $I$ is not an even partition, then we know that one of the two conditions hold: (i) $\sum_{i \in I} s_i \leq S - 1 \Rightarrow \sum_{i \in I} v_i \leq 1 - \frac{1}{S}$, or (ii) $\sum_{i \in I} s_i \geq S + 1 \Rightarrow \sum_{i \in I} v_i \geq 1 + \frac{1}{S}$, because the original numbers $s_i$ are integers. Consider these two cases.

If $\sum_{i \in I} s_i \geq S + 1$, then

$$\mathbf{t} < \prod_{i \in I} 2^{-v_i + 2^{-4b}} = 2^{\sum_{i \in I}(-v_i + 2^{-4b})} \leq 2^{\frac{m}{2^{4b}} - (1 + \frac{1}{S})} \leq 2^{-1 - (\frac{1}{2^b} - \frac{1}{2^{3b}})} = l,$$

by using $S \leq 2^b$ and $m \leq b < 2^b$. On the other hand, if $\sum_{i \in I} s_i \leq S - 1$, then

$$\mathbf{t} \geq \prod_{i \in I} 2^{-v_i} = 2^{-\sum_{i \in I} v_i} \geq 2^{-(1-\frac{1}{S})} = 2^{-1+\frac{1}{S}} \geq 2^{-1+\frac{1}{2^b}} = u.$$

Now suppose $I'$ is an even partition. Then we know that the corresponding $\mathbf{t}'$ is

$$2^{-1} \leq \mathbf{t}' < \prod_{i \in I'} 2^{-v_i + 2^{-4b}} = 2^{\sum_{i \in I'}(-v_i + 2^{-4b})} \leq 2^{-1+\frac{1}{2^{3b}}} = a.$$

The distance of $\mathbf{t}'$ to $2^{-1}$ is always less than the distance of $\mathbf{t}$ of a non even partition plus a gap of $2^{-(3b+2)}$:

$$|\mathbf{t}' - 2^{-1}| + 2^{-(3b+2)} \leq a - 2^{-1} + 2^{-(3b+2)} < \min\{u - 2^{-1}, 2^{-1} - l\} \leq |\mathbf{t} - 2^{-1}|, \quad (6)$$

which is proved by analyzing the two elements of the minimization. The first term holds because

$$a + 2^{-(3b+2)} - 2^{-1} < a \cdot 2^{\frac{1}{2^{2b}}} - 2^{-1} = 2^{-1 + \frac{2^{-b} + 2^{-2b}}{2^b}} - 2^{-1} < 2^{-1+\frac{1}{2^b}} - 2^{-1} = u - 2^{-1}.$$

The second comes from the fact that the function $h(b) = a + l + 2^{-(3b+2)} = 2^{-1+\frac{1}{2^{3b}}} + 2^{-1-(\frac{1}{2^b} - \frac{1}{2^{3b}})} + 2^{-(3b+2)}$ is less than 1 for $b = 1, 2$ (by inspection), it is a monotonic increasing function for $b \geq 2$ (the derivative is always positive), and it has $\lim_{b \to \infty} h(b) = 1$. Hence, we conclude that $h(b) < 1$, which implies

$$a + l + 2^{-(3b+2)} < 1 \iff a - 2^{-1} + 2^{-(3b+2)} < 2^{-1} - l.$$

This concludes that there is a gap of at least $2^{-(3b+2)}$ between the worst value of $\mathbf{t}'$ (relative to an even partition) and the best value of $\mathbf{t}$ (relative to a non even partition), which will be used next to specify the threshold of the MAP problem. Now, set up $\mathbf{X}$ to be the MAP variables and $\mathbf{E} = \mathbf{e}$ and $Y_m = \neg y_m$ to be the evidence, so as we verify if

$$\max_{\mathbf{x} \in \Omega_\mathbf{X}} p(\mathbf{x}, \mathbf{e}^\mathbf{T}, \neg y_m^T) > r = c \cdot \frac{1}{2^m}, \quad (7)$$

where $c$ equals $a' \cdot (1 - a')$, and $a'$ equals $a$ evaluated up to $3b + 2$ bits and rounded up, which implies that $2^{-1} < a \leq a' < a + 2^{-(3b+2)}$.[7] By Eq. (6), $a'$ is closer to one half than any $\mathbf{t}$ of a non even partition, so the value $c$ is certainly greater than any value that would be obtained by a non even partition. On the other hand, $a'$ is farther from $2^{-1}$ than $a$, so we can conclude that

$$\mathbf{t} \cdot (1 - \mathbf{t}) < c \leq a \cdot (1 - a) < \mathbf{t}' \cdot (1 - \mathbf{t}')$$

for any $\mathbf{t}$ corresponding to a non-even partition and any $\mathbf{t}'$ of an even partition. Thus, a solution of the MAP problem obtains $p(\mathbf{x}, \mathbf{e}^\mathbf{T}, \neg y_m^Y) > r$ if and only there is an even partition. $\square$

**Corollary 7** *Decision-MAP is NP-complete when restricted to binary polytrees.*

**Proof** It follows directly from Theorems 5 and 6. $\square$

The next theorem shows that the problem remains hard even in trees. The tree used for the proof is probably the simplest practical tree: a Naive Bayes structure [16], where there is a node called "class" with direct children called "features". These features are independent of each other given the class. The theorem can be easily formulated using other trees, such as a Hidden Markov Model topology [17], by simply replicating the node corresponding to the class. One strong characteristic of the following result is that the reduction is done using the maximum-satisfiability problem. The same problem was employed before to show that MAP is hard in polytrees [1]. Hence, we show here that the inapproximability results for MAP [1] when the maximum cardinality is not bounded extend to the subcase of trees.

---

[7]The conditional version of Decision-MAP could be used in the reduction too by including the term $\frac{1}{p(\mathbf{e}^\mathbf{T}, \neg y_m^T)}$ in $r$, which can be computed in polynomial time by a BN propagation in polytrees [7], and it does not depend on the choice $\mathbf{x}$.

**Theorem 8** *Decision-MAP-∞-w is NP-hard even if $w = 1$ and the network topology is as simple as a Naive Bayes structure.*

**Proof** We use a reduction from MAX-2-SAT. Let $X_1 \ldots, X_m$ be variables of a SAT problem with clauses $C_1, \ldots, C_{m'}$ written in 2CNF, that is, each clause is composed of a disjunction of two literals. Each literal belongs to $\Omega_{X_j} = \{x_j, \neg x_j\}$ for a given $j$. Without loss of generality, we assume that each clause involves exactly two distinct variables of $X_1 \ldots, X_m$. Let $b > 0$ be the number of bits to specify the MAX-2-SAT problem.

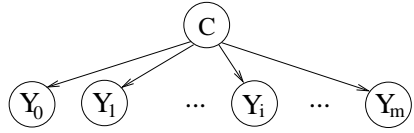

Figure 2: Network structure for the proof of Theorem 8.

Take a Naive Bayes shaped network. Let $C$ be the root of the Naive Bayes structure and $Y_1, \ldots, Y_m$ the binary features (as shown in Figure 2) such that $\Omega_{Y_j} = \{y_j^T, y_j^F\}$ for every $j$. Define the variable $C$ to have $2m'$ states and uniform prior, that is, $p(c) = \frac{1}{2m'}$ for every $c \in \Omega_C$, where $\Omega_C$ equals to $\{c_{1L}, c_{1R}, c_{2L}, c_{2R}, \ldots, c_{m'L}, c_{m'R}\}$. Denote by $L_i$ the literal of clause $C_i$ with the smallest index, and by $R_i$ the literal with the greatest index. Define the conditional probability functions of each $Y_j$ given $C$ as follows:

$$p(y_j^T|c_{iL}) = p(y_j^T|c_{iR}) = \frac{1}{2} \quad \text{if } L_i, R_i \notin \Omega_{X_j}, \text{ that is, } X_j \text{ does not appear in clause } C_i.$$

$$p(y_j^T|c_{iR}) = 1 \quad \text{if } (x_j = R_i) \vee (\neg x_j = L_i).$$
$$p(y_j^T|c_{iR}) = 0 \quad \text{if } (x_j = L_i) \vee (\neg x_j = R_i).$$

$$p(y_j^T|c_{iL}) = 1 \quad \text{if } x_j = L_i.$$
$$p(y_j^T|c_{iL}) = 0 \quad \text{if } \neg x_j = L_i.$$
$$p(y_j^T|c_{iL}) = \frac{1}{2} \quad \text{if } R_i \in \Omega_{X_j}.$$

Define $Y_0$ as an extra feature such that $p(y_0^T|c_{iL}) = 1$ and $p(y_0^T|c_{iR}) = 1/2$ for every $i$. The probability values for $y_j^F$ complement these numbers in order to sum one, that is, $p(y_j^F|c) = 1 - p(y_j^T|c)$ for every $c \in \Omega_C$. Now,

$$\max_{y_0 \ldots y_m} p(y_0, y_1, \ldots, y_m) = \max_{y_1, \ldots y_m} p(y_0^T, y_1, \ldots y_m),$$

because the vector $p(y_0^T|C)$ (note that the vector ranges over the states of $C$, and $y_0^T$ is fixed) pareto-dominates the vector $p(y_0^F|C)$ by construction, that is, $p(y_0^T|c) \geq p(y_0^F|c)$ for every $c \in \Omega_C$ (more details on pareto sets will follow in Section 4, but here it is enough to see that there is no reason to choose $y_0^F$ in place of $y_0^T$ as the probability value of the latter is always greater than that of the former for every given $c$).[8] It is clear that the transformation is polynomial in $b$, as the network has $m+1$ nodes, with at most $2m'$ states (both $m$ and $m'$ are $\Omega(b)$), and the probability values are always 0, 1/2 or 1.

By simple manipulations, we have (for any given $y_1, \ldots, y_m$):

$$p(y_0^T, y_1, \ldots y_m) = \sum_i (p(c_{iL}) \prod_j p(y_j|c_{iL}) + p(c_{iR}) \prod_j p(y_j|c_{iR})) = \frac{1}{2m'} \sum_i (\prod_j p(y_j|c_{iL}) + \prod_j p(y_j|c_{iR}))$$

---

[8] Another approach would force $Y_0 = y_0^T$ as evidence instead of using this pareto argument, which would also suffice to prove the theorem.

$$= \frac{1}{2m'} \frac{1}{2^{m-2}} \sum_i \left( p(y_{j_{iL}}|c_{iL}) p(y_{j_{iR}}|c_{iL}) p(y_0^T|c_{iL}) + p(y_{j_{iL}}|c_{iR}) p(y_{j_{iR}}|c_{iR}) p(y_0^T|c_{iR}) \right),$$

where $j_{iL}$ and $j_{iR}$, with $j_{iL} < j_{iR}$, are the indices of the two variables that happen in clause $C_i$ (the probability of all other variables $Y_j$ that appeared in the product have led to the fraction $\frac{1}{2}$ because they do not happen in $C_i$ and hence disappeared to form the constant $\frac{1}{2^{m-2}}$ that has been put outside the summation). Yet by construction, we continue with

$$\begin{aligned} p(y_0^T, y_1, \ldots y_m) &= \frac{1}{2m'} \frac{1}{2^{m-2}} \sum_i \left( p(y_{j_{iL}}|c_{iL}) \frac{1}{2} + p(y_{j_{iL}}|c_{iR}) p(y_{j_{iR}}|c_{iR}) \frac{1}{2} \right) \\ &= \frac{1}{2m'} \frac{1}{2^{m-1}} \sum_i \left( p(y_{j_{iL}}|c_{iL}) + p(y_{j_{iL}}|c_{iR}) p(y_{j_{iR}}|c_{iR}) \right). \end{aligned}$$

Now note that $p(y_{j_{iL}}|c_{iL})$ and $p(y_{j_{iL}}|c_{iR})$ are mutually exclusive, and that $p(y_{j_{iL}}|c_{iL}) = 1$ if and only if $L_i$ satisfies clause $C_i$. In this case, $p(y_{j_{iL}}|c_{iR}) = 0$, and the sum $p(y_{j_{iL}}|c_{iL}) + p(y_{j_{iL}}|c_{iR}) p(y_{j_{iR}}|c_{iR})$ equals to 1. On the other hand, if $L_i$ does not make clause $C_i$ satisfiable, then $p(y_{j_{iL}}|c_{iL}) + p(y_{j_{iL}}|c_{iR}) p(y_{j_{iR}}|c_{iR}) = p(y_{j_{iR}}|c_{iR})$, that is, it becomes one if and only if $R_i$ makes clause $C_i$ satisfiable. Because we sum over all the clauses, $p(y_0^T, y_1, \ldots y_m) = \frac{k}{2^m m'} \iff k$ clauses are satisfiable in the 2CNF formula. Hence, solving the optimization $\max_{y_0 \ldots y_m} p(y_0, y_1, \ldots, y_m)$ is the same as solving MAX-2-SAT. Because the optimization versions agree as described, the reduction of the decision version follows too. □

We show next a stronger inapproximability result than those previously stated in the literature, because we make use of trees while previous results make use of polytrees (or more sophisticated topologies). Recall that an approximation algorithm for a maximization problem where the exact maximum value is $M > 0$ is said to achieve a ratio $r^0 > 1$ from the optimal if the resulting value is guaranteed to be greater than or equal to $\frac{M}{r^0}$. We demonstrate that approximating Decision-MAP is NP-hard even if the network topology is as simple as a tree. This leaves no hope of approximating MAP in polynomial time when the number of states per variable is not bounded.

**Theorem 9** *It is NP-hard to approximate Decision-MAP-$\infty$-w, with $w = 1$, to any ratio $r^0 = size(\mathcal{N})^\epsilon$ for fixed $\epsilon$, as well as to any ratio $r^0 = 2^{size(\mathcal{N})^\varepsilon}$, for fixed $0 \leq \varepsilon < 1$.*

**Proof** We show that it is possible to reduce MAX-2-SAT to the approximate version of Decision-MAP-$\infty$-1 in polynomial time and space in size($\mathcal{N}$). The idea is similar to the repeated construction used in [1]. We build $q$ copies of the network of Theorem 8 (superscripts are added to the variables to distinguish the copies as follows: the nodes of the $t$-th copy are named $C^t, Y_0^t, \ldots, Y_m^t$) and link them by a common binary parent $D$ of all the $C^t$ nodes (as shown in Fig.3), with states $\{d^T, d^F\}$. We define $p(d^T) = 1$ and $p(c^t|d^T)$ remains uniform as before, for every node $C^t$.

By construction, we have

$$p(y_0^1, \ldots y_m^1, \ldots, y_0^q, \ldots y_m^q) = \sum_d \prod_{t=1}^q p(y_0^t, y_1^t, \ldots y_m^t|d) p(d) = \prod_{t=1}^q p(y_0^t, y_1^t, \ldots y_m^t|d^T),$$

and hence each copy has independent computations given $d^T$. Using the same argument as in Theorem 8 for each copy, we obtain

$$p(y_0^1, \ldots y_m^1, \ldots, y_0^q, \ldots y_m^q) = \prod_{t=1}^q p(y_0^t, \ldots y_m^t) = \prod_{t=1}^q \frac{k}{2^m m'} = \left( \frac{k}{2^m m'} \right)^q$$

if and only if $k$ clauses are satisfiable in the MAX-2-SAT problem. Suppose we want to decide if at least $1 < k' \leq m'$ clauses are satisfiable (the restriction of $k' > 1$ does not lose generality). Using the approximation over this new network with ratio $r^0$, if at least $k'$ clauses are satisfiable, then we must have

$$p(y_0^1, \ldots y_m^1, \ldots, y_0^q, \ldots y_m^q) \geq \frac{1}{r^0} \left( \frac{k'}{2^m m'} \right)^q.$$

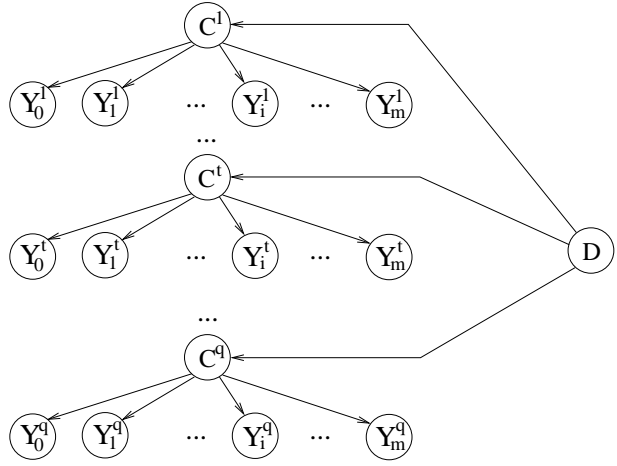

Figure 3: Network structure for the proof of Theorem 9.

On the other hand, if it is not possible to satisfy $k'$ clauses, then we know that

$$p(y_0^1,\ldots y_m^1,\ldots,y_0^q,\ldots y_m^q) \leq \left(\frac{k'-1}{2^m m'}\right)^q.$$

Now we need to show that it is possible to pick $q$ such that $\left(\frac{k'-1}{2^m m'}\right)^q < \frac{1}{r^0}\left(\frac{k'}{2^m m'}\right)^q$ and such that $q$ is polynomially bounded.

$$\left(\frac{k'-1}{2^m m'}\right)^q < \frac{1}{r^0}\left(\frac{k'}{2^m m'}\right)^q \iff r^0 < \left(\frac{k'}{k'-1}\right)^q \iff q > \frac{\log(r^0)}{\log\left(\frac{k'}{k'-1}\right)}.$$

Now, by Taylor expansion, $\log\left(\frac{k'}{k'-1}\right) \geq \frac{1}{k'}$ (for any $k' > 1$) and thus

$$q > k'\log(r^0) \implies q > \frac{\log(r^0)}{\log\left(\frac{k'}{k'-1}\right)} \implies \left(\frac{k'-1}{2^m m'}\right)^q < \frac{1}{r^0}\left(\frac{k'}{2^m m'}\right)^q.$$

The proof concludes by choosing the appropriate $q$. In the case of $r^0 = \text{size}(\mathcal{N})^\epsilon$, we can choose a $q \geq 3$ such that

$$\frac{q}{\log(q)} > \epsilon k'(1 + \log(f'(b))) \implies q > \epsilon k'(\log(q) + \log(f'(b))\log(q)) \implies q > \epsilon k'(\log(q) + \log(f'(b)))$$

$$\implies q > \epsilon k' \log(q \cdot f'(b)) \implies q > k' \log((q \cdot f'(b))^\epsilon) \implies q > k' \log(\text{size}(\mathcal{N})^\epsilon) \implies q > k' \log(r^0),$$

where $f'$ is a polynomial that bounds the size of each copy, given by the construction in Theorem 8, and hence $q$ can be chosen such as to ensure the polynomial transformation in the input size ($\frac{q}{\log(q)}$ is monotonically increasing for $q \geq 3$).

In the case of $r^0 = 2^{\text{size}(\mathcal{N})^\varepsilon}$, then we can choose a $q$ such that

$$q > (\log(2)k'f'(b)^\varepsilon)^{\frac{1}{1-\varepsilon}} \implies q^{1-\varepsilon} > \log(2)k'f'(b)^\varepsilon \implies q > \log(2)k'q^\varepsilon \cdot f'(b)^\varepsilon$$

$$\implies q > \log(2)k'(q \cdot f'(b))^\varepsilon \implies q > k' \log(2^{(q \cdot f'(b))^\varepsilon}) \implies q > k' \log(2^{\text{size}(\mathcal{N})^\varepsilon}) \implies q > k' \log(r^0),$$

and again the choice of $q$ is polynomial in the input size. □

To conclude this section about hardness results, we show that MAP remains hard even in a tree with bounded maximum cardinality. However, we demonstrate in the next sections that many practical problems can be solved exactly, and that fully polynomial approximation schemes are possible when cardinality and treewidth are bounded.

**Theorem 10** *Decision-MAP-z-w is NP-hard even if $z = 5$ and $w = 1$ and the network topology is as simple as a Hidden Markov Model structure.*

**Proof** This proof uses the same problem of the proof of Theorem 6 to perform the reduction, as well as the idea of approximating exponentials to guarantee the polynomial time reduction and the rationality of the numbers. Thus, hardness is shown using a reduction from *partition* problem, which is NP-hard [3] and can be stated as follows: *given a set of $m$ positive integers $s_1, \ldots, s_m$, is there a set $I \subset A = \{1, \ldots, m\}$ such that $\sum_{i \in I} s_i = \sum_{i \in A \setminus I} s_i$?* All the input is encoded using $b > 0$ bits. Furthermore, we assume that $S = \frac{1}{2} \sum_{i \in A} s_i \geq 2$ and we call *even partition* a subset $I \subset A$ that achieves $\sum_{i \in I} s_i = S$. To solve *partition*, we consider the rescaled problem (dividing every element by $S$), so as $v_i = \frac{s_i}{S} \leq 2$ are the elements and we look for a partition with sum equals to 1 (altogether the elements sum 2).

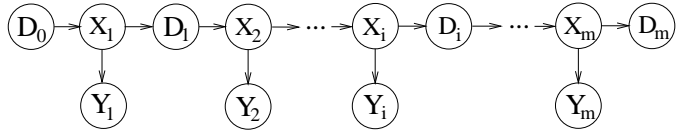

Figure 4: Network structure for the proof of Theorem 10.

We construct (in polynomial time) a tree with $3m + 1$ nodes: $\mathbf{X} = \{X_1, \ldots, X_m\}$, $\mathbf{Y} = \{Y_1, \ldots, Y_m\}$ and $\mathbf{D} = \{D_0, D_1, \ldots, D_m\}$, such that $\Omega_{X_i} = \{x_{i1}, x_{i2}, x_{i3}, x_{i4}, x_{i5}\}$ has 5 states, $\Omega_{Y_i} = \{y_i^T, y_i^F\}$ is binary, and $\Omega_{D_i} = \{d_i^T, d_i^F, d_i^*\}$ is ternary. The structure of the network is presented in Figure 4. The probability functions are defined by Table 1 (except for the probability function of $D_0$, which is uniform).

First, we show that $p(\mathbf{y}) = p(y_1, \ldots, y_m) = \frac{1}{2^m}$ for any configuration $\mathbf{y} \in \Omega_\mathbf{Y}$. By construction, we have $p(y_i | d_{i-1}) = \sum_{x_i} p(y_i | x_i) p(x_i | d_{i-1}) = \frac{1}{2}$ for any value of $y_i, d_{i-1}$. Now,

$$p(y_1, \ldots, y_m) = \sum_{d_{m-1}} p(y_m | d_{m-1}) p(d_{m-1}, y_1, \ldots, y_{m-1}) = \frac{1}{2} \sum_{d_{m-1}} p(d_{m-1}, y_1, \ldots, y_{m-1}) = \frac{1}{2} p(y_1, \ldots, y_{m-1}).$$

Applying the same idea successively, we obtain $p(y_1, \ldots, y_m) = \frac{1}{2^m}$. Using a similar procedure, we can obtain the values for $p(d_m^T, \mathbf{y})$ and $p(d_m^F, \mathbf{y})$ as follows:

$$p(d_m^T, \mathbf{y}) = p(d_m^T, y_1 \ldots, y_m) = \sum_{x_m, d_{m-1}} p(d_m^T, y_m | x_m) p(x_m | d_{m-1}) p(d_{m-1}, y_1, \ldots, y_{m-1})$$

$$= \begin{cases} t_i \cdot \frac{1}{2} \cdot p(d_{m-1}^T, y_1, \ldots, y_{m-1}) & \text{if } y_m = y_m^T, \\ 1 \cdot \frac{1}{2} \cdot p(d_{m-1}^T, y_1, \ldots, y_{m-1}) & \text{if } y_m = y_m^F, \end{cases}$$

and applying successively again, we get $p(d_m^T, y_1 \ldots, y_m) = \frac{2^{-m}}{3} \prod_{i \in I} t_i$, where where $I \subset A$ is the indicator of the elements such that $Y_i$ is at the state $y_i^T$ (the ratio $\frac{1}{3}$ comes from the uniform $p(D_0)$).

$$p(d_m^F, \mathbf{y}) = p(d_m^F, y_1 \ldots, y_m) = \sum_{x_m, d_{m-1}} p(d_m^F, y_m | x_m) p(x_m | d_{m-1}) p(d_{m-1}, y_1, \ldots, y_{m-1})$$

$$= \begin{cases} 1 \cdot \frac{1}{2} \cdot p(d_{m-1}^F, y_1, \ldots, y_{m-1}) & \text{if } y_m = y_m^T, \\ t_i \cdot \frac{1}{2} \cdot p(d_{m-1}^F, y_1, \ldots, y_{m-1}) & \text{if } y_m = y_m^F, \end{cases}$$

and hence $p(d_m^F, y_1 \ldots, y_m) = \frac{2^{-m}}{3} \prod_{i \in A \setminus I} t_i$. Because of that, we have

$$\max_\mathbf{y} p(d_m^*, \mathbf{y}) = \max_\mathbf{y} \left( p(\mathbf{y}) - p(d_m^T, \mathbf{y}) - p(d_m^F, \mathbf{y}) \right) = \frac{1}{2^m} - \min_\mathbf{y} \left( p(d_m^T, \mathbf{y}) + p(d_m^F, \mathbf{y}) \right)$$

$$= \frac{1}{2^m} - \min_\mathbf{y} \left( \frac{2^{-m}}{3} \prod_{i \in I} t_i + \frac{2^{-m}}{3} \prod_{i \in A \setminus I} t_i \right) = \frac{1}{2^m} \left( 1 - \frac{1}{3} \min_\mathbf{y} \left( \prod_{i \in I} t_i + \prod_{i \in A \setminus I} t_i \right) \right).$$

Table 1: Probability functions used in the proof of Theorem 10.

| $p(Y_i\|X_i)$ | $x_{i1}$ | $x_{i2}$ | $x_{i3}$ | $x_{i4}$ | $x_{i5}$ |
|---|---|---|---|---|---|
| $y_i$ | 1 | 1 | 0 | 0 | 1/2 |
| $\neg y_i$ | 0 | 0 | 1 | 1 | 1/2 |

| $p(X_i\|D_{i-1})$ | $d^T_{i-1}$ | $d^F_{i-1}$ | $d^*_{i-1}$ |
|---|---|---|---|
| $x_{i1}$ | 1/2 | 0 | 0 |
| $x_{i2}$ | 0 | 1/2 | 0 |
| $x_{i3}$ | 0 | 1/2 | 0 |
| $x_{i4}$ | 1/2 | 0 | 0 |
| $x_{i5}$ | 0 | 0 | 1 |

| $p(D_i\|X_i)$ | $x_{i1}$ | $x_{i2}$ | $x_{i3}$ | $x_{i4}$ | $x_{i5}$ |
|---|---|---|---|---|---|
| $d^T_i$ | $t_i$ | 0 | 0 | 1 | 0 |
| $d^F_i$ | 0 | 1 | $t_i$ | 0 | 0 |
| $d^*_i$ | $1-t_i$ | 0 | $1-t_i$ | 0 | 1 |

Table 2: Joint probability function of $Y_i, D_i$ given $X_i$ used in the proof of Theorem 10.

| $p(Y_i, D_i\|X_i)$ | $x_{i1}$ | $x_{i2}$ | $x_{i3}$ | $x_{i4}$ | $x_{i5}$ |
|---|---|---|---|---|---|
| $y^T_i, d^T_i$ | $t_i$ | 0 | 0 | 0 | 0 |
| $y^T_i, d^F_i$ | 0 | 1 | 0 | 0 | 0 |
| $y^T_i, d^*_i$ | $1-t_i$ | 0 | 0 | 0 | 1/2 |
| $y^F_i, d^T_i$ | 0 | 0 | 0 | 1 | 0 |
| $y^F_i, d^F_i$ | 0 | 0 | $t_i$ | 0 | 0 |
| $y^F_i, d^*_i$ | 0 | 0 | $1-t_i$ | 0 | 1/2 |

For the sake of simplicity, consider first that $t_i = 2^{-v_i}$. Then, the function $\prod_{i\in I} t_i + \prod_{i\in A\setminus I} t_i = 2^{-\sum_{i\in I} v_i} + 2^{-\sum_{i\in A\setminus I} v_i}$ is convex and achieves its minimum when $2^{-\sum_{i\in I} v_i} = 2^{-\sum_{i\in A\setminus I} v_i} \iff \sum_{i\in I} v_i = \sum_{i\in A\setminus I} v_i = 1$. Thus, using $\mathbf{Y}$ as the MAP variables and $d^*_m$ as the evidence, we obtain $\max_{\mathbf{y}} p(d^*_m, \mathbf{y}) = \frac{2}{3}\frac{1}{2^m}$ if and only if there is an even partition. This is still flaw in one respect: the specification of the probability functions depends on computing the values $t_i$, for each $i \in A$, which can be done only to a certain precision (we can only use a number of places that is polynomial in $b$).

Let $t_i$ be $2^{-v_i}$ computed with $6b+3$ bits of precision and rounded up (if necessary). This implies in $2^{-v_i} \leq t_i < 2^{-v_i+2^{-6b}}$ (by Corollary 16). Suppose first that $I \subset A$ is an even partition. Then,

$$\prod_{i\in I} t_i + \prod_{i\in A\setminus I} t_i < 2^{\sum_{i\in I}(-v_i + 2^{-6b})} + 2^{\sum_{i\in A\setminus I}(-v_i + 2^{-6b})} \leq 2^{-\sum_{i\in I} v_i + m 2^{-6b}} + 2^{-\sum_{i\in A\setminus I} v_i + m 2^{-6b}}$$

$$\leq 2^{-\sum_{i\in I} v_i + 2^{-5b}} + 2^{-\sum_{i\in A\setminus I} v_i + 2^{-5b}} = 2^{-1+2^{-5b}} + 2^{-1+2^{-5b}} = 2^{2^{-5b}},$$

and

$$\frac{1}{2^m}\left(1 - \frac{1}{3}\min_{\mathbf{y}}\left(\prod_{i\in I} t_i + \prod_{i\in A\setminus I} t_i\right)\right) > \frac{1}{2^m}\left(1 - \frac{1}{3} 2^{2^{-5b}}\right).$$

Take now the case where $I \subset A$ is not an even partition. Without loss of generality, suppose that $\sum_{i\in I} s_i = S - l$, with $0 < l \leq S$ an integer. This implies in $\sum_{i\in I} v_i = 1 - l/S$ and $\sum_{i\in A\setminus I} v_i = 1 + l/S$. In this case, we have (because $t_i$s were rounded up, if needed):

$$\prod_{i\in I} t_i + \prod_{i\in A\setminus I} t_i \geq 2^{\sum_{i\in I} -v_i} + 2^{\sum_{i\in A\setminus I} -v_i} \geq 2^{-1+l/S} + 2^{-1-l/S}.$$

We show that $2^{-1+l/S} + 2^{-1-l/S} \geq 2^{2^{-4b}}$. Note that $2^{-1+l/S} + 2^{-1-l/S}$ is a convex function on $l$ and achieves its minimum when $l$ is minimized. Thus, it is enough to show that $2^{-1+1/S} + 2^{-1-1/S} \geq 2^{2^{-4b}}$ (because $l$ is a positive integer). Let $x = 1/S$. Note that $2^{-b} \leq x \leq 1/2$, because $2 \leq S \leq 2^b$. With some manipulations we have

$$2^{-1+x} + 2^{-1-x} \geq 2^{2^{-4b}} \iff 1 + 2^{2x} \geq 2^{2^{-4b}+1+x} \iff \log_2(1 + 2^{2x}) \geq 2^{-4b} + 1 + x,$$

and thus it is enough to have $\log_2(1 + 2^{2x}) - x^4 - 1 - x \geq 0$ (because $x \geq 2^{-b}$), which follows by Lemma 17 (as $x \leq 1/2$).

At this point we know that an even partition leads to a value of $\prod_{i \in I} t_i + \prod_{i \in A \setminus I} t_i$ that is less than $2^{2^{-5b}}$, while a non even partition to a value that is greater than $2^{2^{-4b}}$. Now, we pick a threshold $r = \frac{1}{2^m}(1 - \frac{a}{3})$, where $a$ equals to $2^{2^{-5b}}$ computed with $5b + 3$ bits of precision and rounded up (if necessary) to decide if the partition is even. In this way, $2^{2^{-5b}} \leq a < 2^{2^{-5b} + 2^{-5b}} = 2^{2^{-5b+1}} \leq 2^{2^{-4b}}$, and thus $r$ separates the cases with even and non even partitions:

$$p(d_m^*, \mathbf{y}^{\text{non-even}}) \leq \frac{1}{2^m}\left(1 - \frac{1}{3}2^{2^{-4b}}\right) \leq r = \frac{1}{2^m}\left(1 - \frac{a}{3}\right) \leq \frac{1}{2^m}\left(1 - \frac{1}{3}2^{2^{-5b}}\right) < p(d_m^*, \mathbf{y}^{\text{even}}). \square$$

**Corollary 11** *Decision-MAP is NP-complete when the graph is restricted to a tree and variables have bounded cardinality.*

**Proof** It follows directly from Theorems 5 and 10. $\square$

### 4. A new algorithm for MAP

Despite the "negative" complexity results of previous section, this section describes a considerably fast exact algorithm for MAP, which is later extended to run in an approximate way. Variables that are part of the MAP query will be denote $\mathcal{X}^{\text{map}} \subseteq \mathcal{X} \setminus \mathbf{E}$ throughout the section. The basis to solve the problem is to compute $p(\mathbf{x}^{\text{map}}, \mathbf{e})$ using Eq. (2) for every possible $\mathbf{x}^{\text{map}} \in \Omega_{\mathcal{X}^{\text{map}}}$, but that would need $z(\mathcal{X}^{\text{map}}) = \prod_{X_i \in \mathcal{X}^{\text{map}}} z_i$ evaluations of Eq. (2), which is the same as a brute-force approach. Another approach is to propagate the information just as in the BU query, but considering many probability functions for different instantiations of $\mathbf{x}^{\text{map}}$ in every possible way, yet somehow locally. We will use the notation $p_{\mathbf{x}^{\text{map}}}(\mathbf{u}|\mathbf{v}) = p(\mathbf{x}^{\text{map}}, \mathbf{u}|\mathbf{v})$, where $\mathbf{x}^{\text{map}} \in \Omega_{\mathbf{X}^{\text{map}}}$ (with $\mathbf{X}^{\text{map}} \subseteq \mathcal{X}^{\text{map}}$, being clear later from the context), $\mathbf{u} \in \Omega_{\mathbf{U}}$ (with $\mathbf{U} \subseteq \mathcal{X} \setminus \mathcal{X}^{\text{map}}$), $\mathbf{v} \in \Omega_{\mathbf{V}}$ (with $\mathbf{V} \subseteq \mathcal{X} \setminus (\mathbf{U} \cup \mathbf{X}^{\text{map}}))$, and hence we have distinct functions $p_{\mathbf{x}^{\text{map}}}: \Omega_{\mathbf{U}} \times \Omega_{\mathbf{V}} \to \mathcal{Q}$ for distinct instantiations $\mathbf{x}^{\text{map}}$. We assume further that $\mathbf{u}, \mathbf{v}$ always respect the instantiation of the evidence $\mathbf{e}$. For convenience, $p_{\mathbf{x}^{\text{map}}}(\mathbf{U}|\mathbf{V})$, or simply $p_{\mathbf{x}^{\text{map}}}$, may be specified by the vector $[p_{\mathbf{x}^{\text{map}}}(\mathbf{u}|\mathbf{v})]_{\forall (\mathbf{u} \in \Omega_{\mathbf{U}}, \mathbf{v} \in \Omega_{\mathbf{V}}): (\mathbf{u}, \mathbf{v}) \text{ respect } \mathbf{e}}$ in $\mathcal{Q}^d$, where $d = z((\mathbf{U} \cup \mathbf{V}) \setminus \mathbf{E})$, as the part of $\mathbf{u}, \mathbf{v}$ corresponding to $\mathbf{e}$ is fixed to the observed states.

Let $(\mathbf{C}, T)$ be a tree decomposition of the network. The main idea of the algorithm is quite simple: we propagate through the tree only the functions $p_{\mathbf{x}^{\text{map}}}$ that are not dominated by others, that is, at each step we keep only the pareto set of all $p_{\mathbf{x}^{\text{map}}}$.

**Definition 12** *Let $S$ be a subset of $\mathcal{Q}^q$ (for a fixed dimension $q$), and $a \in S$ and $b \in S$ be two $d$-dimensional vectors of rationals such that $a_i, b_i$ are the $i$-th values of $a, b$, respectively. We say that $a$ dominates $b$, and indicate by $a \succ b$, if $a_i \geq b_i$ for every $1 \leq i \leq d$ and $a_i > b_i$ for at least one $i$. A vector $a \in S$ is called* non-dominated *in $S$ if there is no vector $b \in S$ such that $b \succ a$. A set of non-dominated vectors is a* pareto set.

Note that for each instantiation $\mathbf{x}^{\text{map}}$, the function $p_{\mathbf{x}^{\text{map}}}(\mathbf{U}|\mathbf{V})$ is simply a multi-dimensional vector containing probability values for each instantiation of its arguments $\mathbf{U}$ and $\mathbf{V}$ (except for those in $\mathbf{E}$). In view of Def. 12, two vectors $a = p_{\mathbf{x}_1^{\text{map}}}$ and $b = p_{\mathbf{x}_2^{\text{map}}}$ in the same dimension $d = z((\mathbf{U} \cup \mathbf{V}) \setminus \mathbf{E})$ can be compared to verify if one dominates the other. The algorithm proceeds as follows: Eq. (8) is recursively evaluated, starting from the leaves of the tree. At each step $\mathbf{C}_j$, it uses every combination of vectors $p_{\mathbf{x}_{\mathbf{C}_{j'}}^{\text{map}}}$ in the pareto sets previously obtained at the children nodes $\mathbf{C}_{j'}$ to compute a new pareto set of vectors $p_{\mathbf{x}_{\mathbf{C}_j}^{\text{map}}}$ as the output of the node $\mathbf{C}_j$, which will be propagated to the parent of $\mathbf{C}_j$. The notation $\mathbf{x}_{\mathbf{C}_j}^{\text{map}}$ stands for the queried MAP variables that have already been processed, that is, they appear only in $\mathbf{C}_j$ and/or its descendants and do not appear in $\mathbf{C}_{j_p}$ (the parent of $\mathbf{C}_j$ in $T$).

$$p_{\mathbf{x}_{\mathbf{C}_j}^{\text{map}}}(\mathbf{u}_{\mathbf{C}_j}|\mathbf{v}_{\mathbf{C}_j}) = \sum_{\Omega_{\mathbf{X}_j^{\text{last}} \setminus (\mathbf{E} \cup \mathbf{X}^{\text{map}})}} \prod_{X_i \in \mathbf{X}_j^{\text{proc}}} p(x_i|\pi_{X_i}) \cdot \prod_{\mathbf{C}_{j'} \in \Lambda_{\mathbf{C}_j}} p_{\mathbf{x}_{\mathbf{C}_{j'}}^{\text{map}}}(\mathbf{u}_{\mathbf{C}_{j'}}|\mathbf{v}_{\mathbf{C}_{j'}}). \tag{8}$$

Note that all the elements $x_i$, $\pi_{X_i}$ (for each $X_i$), and $\mathbf{u}_{\mathbf{C}_{j'}}$, $\mathbf{v}_{\mathbf{C}_{j'}}$ (for each $j'$) must agree with the evidence $\mathbf{e}$ (this is done by instantiating them with the value of the evidence from the beginning).

Now, the advantage here is to take into account that only vectors $p_{\mathbf{x}_{\mathbf{C}_j}^{\mathrm{map}}}$ belonging to the pareto set are able to produce the value that maximizes the joint probability of the queried variables. This is proved by the following arguments: take a node $\mathbf{C}_j$ and suppose that there is a child $\mathbf{C}_{j'} \in \Lambda_{\mathbf{C}_j}$ such that $p_{\mathbf{x}_{1,\mathbf{C}_{j'}}^{\mathrm{map}}} \succ p_{\mathbf{x}_{2,\mathbf{C}_{j'}}^{\mathrm{map}}}$ (recall that both $p_{\mathbf{x}_{1,\mathbf{C}_{j'}}^{\mathrm{map}}}$ and $p_{\mathbf{x}_{2,\mathbf{C}_{j'}}^{\mathrm{map}}}$ are in fact numeric vectors in a $z((\mathbf{U}_{\mathbf{C}_{j'}} \cup \mathbf{V}_{\mathbf{C}_{j'}}) \setminus \mathbf{E})$ dimensional space, so they are comparable). Let $p_{\mathbf{x}_{1,\mathbf{C}_j}^{\mathrm{map}}}$ and $p_{\mathbf{x}_{2,\mathbf{C}_j}^{\mathrm{map}}}$ be the vectors obtained by the computations at node $\mathbf{C}_j$ using respectively $p_{\mathbf{x}_{1,\mathbf{C}_{j'}}^{\mathrm{map}}}$ and $p_{\mathbf{x}_{2,\mathbf{C}_{j'}}^{\mathrm{map}}}$ (while all other numbers are somehow fixed). If we replace the vector $p_{\mathbf{x}_{2,\mathbf{C}_{j'}}^{\mathrm{map}}}$ by the vector $p_{\mathbf{x}_{1,\mathbf{C}_{j'}}^{\mathrm{map}}}$, we have that every product and/or summation of Eq. (8) will not decrease, and at least one of them will increase, because Eq. (8) is a sum-product of non-negative numbers. This concludes that $p_{\mathbf{x}_{1,\mathbf{C}_j}^{\mathrm{map}}} \succ p_{\mathbf{x}_{2,\mathbf{C}_j}^{\mathrm{map}}}$. As the final objective $p(\mathbf{x}_{\mathbf{C}_1}^{\mathrm{map}}, \mathbf{e}) = p_{\mathbf{x}_{\mathbf{C}_1}^{\mathrm{map}}}(\mathbf{u}_{\mathbf{C}_1})$ is certainly a non-dominated vector in one dimension (otherwise it would not be a maximum solution), an inductive argument over the tree decomposition suffices to show that dominated vectors may be discarded.

In the worst case, the pareto set will be composed of all the vectors, and the procedure will simply run a sophisticated brute-force approach: all the candidates would be propagated through the nodes of the tree decomposition until the corresponding maximizations are performed. However, the expected number of elements in a pareto set created from random vectors is polynomial [18]. Such attractive situation can be seen in our experiments (Section 5). There is another interesting property of this idea: if MAP variables cut the graph (and the corresponding tree decomposition is built to exploit this situation such that all variables propagated from a tree node to its parent are MAP variables or evidence), then the complexity of solving MAP reduces to the complexity of the subparts of the graph. This happens because the information to be propagated between the separate parts, that is, $p_{\mathbf{x}_{\mathbf{C}_j}^{\mathrm{map}}}(\mathbf{u}_{\mathbf{C}_j}|\mathbf{v}_{\mathbf{C}_j})$, reduces to a single number, because if MAP variables cut the graph (and the tree decomposition is built accordingly) such that $(\mathbf{U}_{\mathbf{C}_j} \cup \mathbf{V}_{\mathbf{C}_j}) \setminus \mathbf{E} = \emptyset$, then $z((\mathbf{U}_{\mathbf{C}_j} \cup \mathbf{V}_{\mathbf{C}_j}) \setminus \mathbf{E}) = z(\emptyset) = 1$ (the pareto set will contain only a vector of size one with the maximum value related to the best configuration for the corresponding $\mathbf{x}^{\mathrm{map}}$), and this is pretty much equivalent to solving the problem separately in the subparts (in a proper order). In other words, the worst-case exponential time of MAP is limited to the number of MAP variables that are "visible" to each other (by visible we mean that there is a path in the subjacent graph between the MAP variables that does not contain any MAP variable or evidence). Although such situation might not be very common in general graphs, it might happen often in trees and polytrees, as any variable that is not a leaf or a root node always cuts the graph (in the case of networks other than trees, a simple transformation with an extra node per child of a MAP variable might be used to avoid connections introduced by the moralization of the graph, and thus keeping the cut induced by the MAP variables in the moralized version). For instance, if the MAP variables are randomly positioned in a tree, then the algorithm will probably run very fast, because the number of visible variables (to each other variable) is likely to be very small. Besides that, if an approximation is enough, then an adapted version of this algorithm runs in worst-case polynomial time, as we see in the sequel.

**Theorem 13** *MAP-z-w has a FPTAS for any fixed $z$ and $w$.*

**Proof** We need to show that, for a given $\varepsilon > 0$, there is an algorithm $A$ that is polynomial in $\mathrm{size}(\mathcal{N})$ and in $\frac{1}{\varepsilon}$ such that the value $p^A(\mathbf{x}^{\mathrm{map}}, \mathbf{e})$ obtained by $A$ is at least $\frac{p(\mathbf{x}_{\mathrm{opt}}^{\mathrm{map}}, \mathbf{e})}{1+\varepsilon}$, where $p(\mathbf{x}_{\mathrm{opt}}^{\mathrm{map}}, \mathbf{e})$ stands for the optimal solution value of MAP-z-w, that is, $\mathbf{x}_{\mathrm{opt}}^{\mathrm{map}} = \mathrm{argmax}_{\mathbf{x}^{\mathrm{map}}} p(\mathbf{x}^{\mathrm{map}}, \mathbf{e})$. We know that $p(\mathbf{x}_{\mathrm{opt}}^{\mathrm{map}}, \mathbf{e}) > 0$ because $\sum_{\mathbf{x}^{\mathrm{map}}} p(\mathbf{x}^{\mathrm{map}}, \mathbf{e}) = p(\mathbf{e}) > 0$ and thus the one achieving the maximum cannot be zero. In fact we have that $1 \geq p(\mathbf{e}) \geq p(\mathbf{x}_{\mathrm{opt}}^{\mathrm{map}}, \mathbf{e}) > 2^{-g(\mathrm{size}(\mathcal{N}))}$, where $g : \mathcal{Q} \to \mathcal{Q}$ is a polynomial function, because $p(\mathbf{x}_{\mathrm{opt}}^{\mathrm{map}}, \mathbf{e})$ is obtained from a sequence of $O(nz^w)$ additions and multiplications over numbers of the input. The same argument holds for every intermediate probability value: we have that $p(\mathbf{u}_{\mathbf{C}_j}|\mathbf{v}_{\mathbf{C}_j}) > 0 \Rightarrow p(\mathbf{u}_{\mathbf{C}_j}|\mathbf{v}_{\mathbf{C}_j}) > 2^{-g(\mathrm{size}(\mathcal{N}))}$, for a given polynomial function $g$. Hence, let $g$ be a polynomial function that satisfies such condition for every number involved in the calculations. Let $(\mathbf{C}, T)$ be the tree decomposition of the network of width $w$, which can be obtained in polynomial time [12], and $w' = w + 1$ be the maximum number of variables in a single node of the decomposition. Recall that each $p_{\mathbf{x}_{\mathbf{C}_j}^{\mathrm{map}}}(\mathbf{U}_{\mathbf{C}_j}|\mathbf{V}_{\mathbf{C}_j})$ is a vector in the dimension $z((\mathbf{U}_{\mathbf{C}_j} \cup \mathbf{V}_{\mathbf{C}_j}) \setminus \mathbf{E})$, so the idea is to show that we can fix an upper bound to the number of vectors of the pareto set containing $p_{\mathbf{x}_{\mathbf{C_j}}^{\text{map}}}(\mathbf{U}_{\mathbf{C}_j}|\mathbf{V}_{\mathbf{C}_j})$. For that purpose and following the ideas of [18], we divide the space into a lattice of hypercubes such that, in each coordinate, the ratio of the largest to the smallest value is $1 + \frac{\varepsilon}{2w'n'}$ (recall that $n' \in O(n)$ is the number of nodes in the tree decomposition), which produces a number of hypercubes bounded by

$$O\left(\left(\frac{\log 2^{g(\text{size}(\mathcal{N}))}}{\frac{\varepsilon}{2w'n'}}\right)^{z(\mathbf{U}_{\mathbf{C}_j} \cup \mathbf{V}_{\mathbf{C}_j})}\right) \in O((2(w+1)n \cdot \frac{g(\text{size}(\mathcal{N}))}{\varepsilon})^{z^{w+1}}), \tag{9}$$

because every $p_{\mathbf{x}_{\mathbf{C_j}}^{\text{map}}}(\mathbf{u}_{\mathbf{C}_j}|\mathbf{v}_{\mathbf{C}_j}) \leq 1$, and the log appears because the lattice is created from 1 to $2^{-g(\text{size}(\mathcal{N}))}$, successively dividing the coordinate by $1 + \frac{\varepsilon}{2w'n'}$ (a bin for the exact zero probability is also allocated). In each hypercube, we keep at most one vector, so Eq. (9) bounds the number of elements in the pareto set that is computed at each node $\mathbf{C}_j$. We call this set a *reduced* pareto set. This procedure is carried out over all the nodes of the tree. Therefore, we have a polynomial time procedure both in size($\mathcal{N}$) and in $\frac{1}{\varepsilon}$, because the total running time is less than Eq. (3) times Eq. (9) raised to 2 (using a binary tree decomposition). It remains to show that the resulting $p^A(\mathbf{x}^{\text{map}}, \mathbf{e})$ is at least $\frac{p(\mathbf{x}_{\text{opt}}^{\text{map}}, \mathbf{e})}{1+\varepsilon}$. Each value $p_{\mathbf{x}_{\mathbf{C_j}}^{\text{map}}}(\mathbf{u}_{\mathbf{C}_j}|\mathbf{v}_{\mathbf{C}_j})$ is obtained from a sum of multiplications, with at most $|\Lambda_{\mathbf{C}_j}| + 1$ terms each. Hence, the approximation satisfies

$$p^A_{\mathbf{x}_{\mathbf{C_j}}^{\text{map}}}(\mathbf{u}_{\mathbf{C}_j}|\mathbf{v}_{\mathbf{C}_j}) > \sum_{\Omega_{\mathbf{X}_j^{\text{last}} \setminus (\mathbf{E} \cup \mathbf{X}^{\text{map}})}} \prod_{X_i \in \mathbf{X}_j^{\text{proc}}} p(x_i|\pi_{X_i}) \cdot \prod_{\mathbf{C}_{j'} \in \Lambda_{\mathbf{C}_j}} \frac{p_{\mathbf{x}_{\mathbf{C}_{j'}}^{\text{map}}}(\mathbf{u}_{\mathbf{C}_{j'}}|\mathbf{v}_{\mathbf{C}_{j'}})}{(1 + \frac{\varepsilon}{2w'n'})^{l_{\mathbf{C}_{j'}}}} > \frac{p_{\mathbf{x}_{\mathbf{C_j}}^{\text{map}}}(\mathbf{u}_{\mathbf{C}_j}|\mathbf{v}_{\mathbf{C}_j})}{(1 + \frac{\varepsilon}{2w'n'})^{l_{\mathbf{C}_j}}},$$

as $l_{\mathbf{C}_j}$, the weight of $\mathbf{C}_j$ (which is the number of variables that appear in nodes of the subtree of $T$ rooted at $\mathbf{C}_j$), is equal to $l_{\mathbf{C}_j} = |\mathbf{C}_j| + \sum_{\mathbf{C}_{j'} \in \Lambda_{\mathbf{C}_j}} l_{\mathbf{C}_{j'}}$. Now taking the root of $T$, we have that $p^A(\mathbf{x}^{\text{map}}, \mathbf{e}) > \frac{p(\mathbf{x}_{\text{opt}}^{\text{map}}, \mathbf{e})}{(1+\frac{\varepsilon}{2w'n'})^{w'n'}}$, as $w'n' \geq l_{\mathbf{C}_1}$ (there are less than $w'$ elements per node $\mathbf{C}_j$). It is important to mention that some intermediate values $p^A_{\mathbf{x}_{\mathbf{C_j}}^{\text{map}}}(\mathbf{u}_{\mathbf{C}_j}|\mathbf{v}_{\mathbf{C}_j})$ can be zero, but one can prove by induction in $T$ that this is not a problem for the approximation, because it only happens if the corresponding exact $p_{\mathbf{x}_{\mathbf{C_j}}^{\text{map}},\text{opt}}(\mathbf{u}_{\mathbf{C}_j}|\mathbf{v}_{\mathbf{C}_j})$ is also zero. The proof is as follows: take a leaf as basis. In this case, $p^A_{\mathbf{x}_{\mathbf{C_j}}^{\text{map}}}(\mathbf{u}_{\mathbf{C}_j}|\mathbf{v}_{\mathbf{C}_j})$ is zero only if parameters $p(x_i|\pi_{X_i})$ of the input turn it into a zero, and the value is precise (it means that it is equal to $p_{\mathbf{x}_{\mathbf{C_j}}^{\text{map}},\text{opt}}(\mathbf{u}_{\mathbf{C}_j}|\mathbf{v}_{\mathbf{C}_j})$). Now take an internal node where $p^A_{\mathbf{x}_{\mathbf{C_j}}^{\text{map}}}(\mathbf{u}_{\mathbf{C}_j}|\mathbf{v}_{\mathbf{C}_j})$ is zero. We have that there is a factor in each term of the summation that is zero (because it is a sum of non-negative numbers). If the zero is at a parameter $p(x_i|\pi_{X_i})$ of the input, then the result of the multiplication is precise (the other factors of the multiplication do not matter). Otherwise, if the zero is at a given $p_{\mathbf{x}_{\mathbf{C}_{j'}}^{\text{map}}}(\mathbf{u}_{\mathbf{C}_{j'}}|\mathbf{v}_{\mathbf{C}_{j'}})$, then by hypothesis of the induction that value is precise and not a result of an approximation, so the same argument holds. This concludes that wherever a zero appears in the computations, it does not interfere in the approximation estimation (in fact, it can be only beneficial), so the computation of each $p^A_{\mathbf{x}_{\mathbf{C_j}}^{\text{map}}}(\mathbf{u}_{\mathbf{C}_j}|\mathbf{v}_{\mathbf{C}_j})$ is still within the $\frac{1}{(1+\frac{\varepsilon}{2w'n'})^{l_{\mathbf{C}_j}}}$ ratio.

Finally, $p^A(\mathbf{x}^{\text{map}}, \mathbf{e}) = p^A_{\mathbf{x}_{\mathbf{C_1}}^{\text{map}}}(\mathbf{u}_{\mathbf{C}_1}) > \frac{p_{\mathbf{x}_{\mathbf{C_1}}^{\text{map}}}(\mathbf{u}_{\mathbf{C}_1})}{(1+\frac{\varepsilon}{2w'n'})^{w'n'}} = \frac{p(\mathbf{x}_{\text{opt}}^{\text{map}}, \mathbf{e})}{(1+\frac{\varepsilon}{2w'n'})^{w'n'}} \geq \frac{p(\mathbf{x}_{\text{opt}}^{\text{map}}, \mathbf{e})}{1+\varepsilon}$, because of the inequality $(1 + \frac{\varepsilon}{r})^r \leq 1 + 2\varepsilon$, which is valid for any $0 \leq \varepsilon \leq 1$ and $r$ a positive integer (the left-hand side is convex in $\varepsilon$). □

It follows from Theorem 13 that MAP-$z$-$w$ has a FPTAS in any network with bounded width and cardinality, including polytrees. At first, this result seems to contradict past results, where it is stated that approximating MAP is hard even in polytrees [1]. But note that we assume a bound for the cardinality of the variables, which is the most common situation in practical BNs, while previous results work with a more general class of networks and do not assume the bound. In many applications the number of states of a variable does not increase with the number of nodes (in fact, it is usually much smaller). Finally, we must point out that the result of Theorem 13 uses a multiplicative approximation error, where the target is to be near the optimal solution by considering a worst-case error based on a small multiplicative factor, that is, if $M$ is the true maximum value of the optimization, we can guarantee to find a solution with value at least $M/r^0$, where $r^0 > 1$ is the approximation ratio. Still, it is possible to derive an additive approximation algorithm, where the goal is to be within an additive term with respect to the optimal value, that is, we would ensure to find a solution with value at least $M - r^1$, with $r^1 > 0$ being the additive approximation error.

Table 3: Average results of runs of the algorithms in many random generated networks where SamIam has solved the problem (many lines are missing because no instance was solved by SamIam in those cases. #Q means number of queries. SS means search space size. Within parenthesis and near to time results are the counts of successful runs of each algorithm, for each network type.

| Net type | #Q | SS | SamIam time(sec) | Approx. time(sec) | Exact time(sec) | Avg. pareto | Avg. dimen. |
|---|---|---|---|---|---|---|---|
| alarm.37.nb.(0-10) | 103 | $2^{2.0}$ | 35.4 | 0.0 (103) | 0.0 (103) | 1.2 | 1.4 |
| insurance.27.nb.(0-10) | 80 | $2^{4.4}$ | 104.6 | 0.0 (80) | 0.0 (80) | 2.5 | 9.9 |
| insurance.27.nb.(10-20) | 18 | $2^{13.9}$ | 1492.3 | 3.9 (18) | 8.9 (18) | 337.8 | 132.2 |
| poly.100.(0-10) | 62 | $2^{2.1}$ | 22.2 | 0.0 (62) | 0.0 (62) | 1.0 | 1.2 |
| poly.100.nb.(0-10) | 48 | $2^{2.7}$ | 1.0 | 0.0 (48) | 0.0 (48) | 1.2 | 1.2 |
| poly.50.(0-10) | 55 | $2^{2.9}$ | 14.5 | 0.0 (55) | 0.0 (55) | 1.2 | 1.3 |
| rand.100.(0-10) | 20 | $2^{2.0}$ | 0.0 | 0.0 (20) | 0.0 (20) | 1.0 | 1.1 |
| rand.100.tw4.(0-10) | 22 | $2^{2.3}$ | 1.1 | 0.0 (22) | 0.0 (22) | 1.4 | 1.1 |
| rand.100.tw8.(0-10) | 23 | $2^{2.3}$ | 90.7 | 0.0 (23) | 0.0 (23) | 1.2 | 1.2 |
| rand.30.(0-10) | 45 | $2^{4.5}$ | 196.4 | 0.0 (45) | 0.0 (45) | 2.3 | 2.4 |
| rand.30.(10-20) | 13 | $2^{12.5}$ | 649.1 | 0.1 (13) | 0.1 (13) | 48.2 | 16.7 |
| rand.30.nb.(0-10) | 21 | $2^{2.5}$ | 9.9 | 0.0 (21) | 0.0 (21) | 1.1 | 1.2 |
| rand.30.nb.(10-20) | 1 | $2^{12.0}$ | 1338.0 | 0.0 (1) | 0.0 (1) | 49.0 | 6.0 |
| rand.50.(0-10) | 21 | $2^{2.2}$ | 37.2 | 0.0 (21) | 0.0 (21) | 1.1 | 1.2 |
| rand.50.nb.(0-10) | 61 | $2^{3.1}$ | 63.7 | 0.0 (61) | 0.0 (61) | 1.3 | 1.3 |
| rand.50.tw4.(0-10) | 27 | $2^{2.6}$ | 82.0 | 0.0 (27) | 0.0 (27) | 1.6 | 1.6 |
| rand.50.tw8.(0-10) | 24 | $2^{3.0}$ | 3.3 | 0.0 (24) | 0.0 (24) | 2.0 | 1.5 |
| rand30iw4.(0-10) | 57 | $2^{5.0}$ | 241.6 | 0.0 (57) | 0.0 (57) | 7.3 | 2.5 |
| rand30iw4.(10-20) | 26 | $2^{14.3}$ | 1245.1 | 0.2 (26) | 0.9 (26) | 101.3 | 6.4 |

Additive approximation algorithms are better than their multiplicative counterpart when the optimal value is large, and worse when it is small. The main idea of the proof is similar to that of Theorem 13, but the lattice is built by dividing the space with hypercubes of uniform length (again based on $\varepsilon$, $w$ and $n$). We have chosen to present the multiplicative version of the proof because it is the most common in the theory of approximation algorithms.

## 5. Experiments and final remarks

We perform experiments with the exact method using the structure of some well known networks and some random generated networks. Tables 3, 4 and 5 show the type of the network (names presented in the first column follow the notation *type.size.subtype.limit*, where *type* is in {alarm,insurance,poly,rand} respectively meaning alarm, insurance, polytree, and random topologies; *size* is half of the number of nodes; *subtype* is in {nb,tw4,tw8} meaning respectively non-binary, tree-width equals to 4, tree-width equals to 8; and *limit* indicates that those tests correspond to problems where log (in base 2) of the search space is within this number), the number of queries, the size of the search space (which is the product of the number of states of all the queried MAP variables), the running time of the SamIam package [19], the running time of this procedure using an additive error of (at most) 1%, the running time of this procedure using the exact method, the average number of elements in the pareto set of each step, and the average dimensionality of the vectors in each step. Near the time results are also presented the number of successful runs of each algorithm. All networks have nodes with at most 5 states. Apart from the first two columns, the other numbers are averages over the queries. Table 3 only displays results where the SamIam package was able to solve the corresponding problem within one hour of computation, while Table 4 only present results where the new exact method solved the corresponding problems. Finally, Table 5 presents results for test instances where the new exact method has failed (and consequently also SamIam has failed, because it has solved only a subset of the instances that were solved by the new exact method).

The number of queries in each line is not a constant because we have not generated the networks with a dimensionality constraint (that is, defining a priori the search space size), but instead the space size was verified after the experiments. Around 1700 tests are conducted. We divided the runs into levels of "hardness" to show the differences in running time. MAP variables were connected to the original network variables (using uniform priors) such that they are always in extreme nodes (roots or leaves), which in general generates hard instances (this is the reason why the number of nodes is twice as many). For example, *Alarm.37* has in fact $37 \times 2 = 74$ nodes, and so on. The search space means the number of BU queries to solve the problem by a brute-force approach. The results show that we can exactly solve MAP in networks of practical size. We have also run the state-of-the-art algorithm of the SamIam system [19]. The cells of the tables marked with a *dashed* indicate that the method was unable to output the answer after one hour of computation for any test case. Otherwise, the number within parenthesis indicates how many problems were solved out of the total. Even if we have used the additive approximation, the approximation algorithm provided results that are also within 1% in the multiplicative sense for 99.8% of the tests (and in the few 0.2% of cases where it has not achieved the 1.01 factor, it was still below a multiplicative approximation factor of 1.03 from the optimal value). Furthermore, the approximation results are in fact exact in 67% of the cases (where we have the exact solution to compare against), and have only 0.09% error (in average) in the remaining cases (much better than the worst error guaranteed by the method). Yet the comparison with SamIam shall be viewed in a broad perspective, because differences in the implementations might affect the results. For instance, the tree decomposition (an NP-hard problem) has not been optimized. Nevertheless, the new algorithm reduces time costs by orders of magnitude.

Table 4: Average results of runs of the algorithms in many random generated networks where the new exact method succeeded to solve the problem. #Q means number of queries. SS means search space size. Within parenthesis and near to time results are the counts of successful runs of each algorithm, for each network type.

| Net type | #Q | SS | SamIam time(sec) | Approx. time(sec) | Exact time(sec) | Avg. pareto | Avg. dimen. |
|---|---|---|---|---|---|---|---|
| alarm.37.nb.(0-10) | 120 | $2^{2.8}$ | 35.4 (103) | 0.0 (120) | 0.0 | 1.5 | 2.1 |
| alarm.37.nb.(10-20) | 40 | $2^{16.0}$ | – | 0.0 (40) | 0.0 | 15.7 | 8.8 |
| alarm.37.nb.(20-40) | 30 | $2^{29.3}$ | – | 0.5 (30) | 2.2 | 182.7 | 12.0 |
| alarm.37.nb.($>40$) | 10 | $2^{48.0}$ | – | 38.0 (10) | 113.7 | 1415.1 | 14.0 |
| insurance.27.nb.(0-10) | 110 | $2^{5.5}$ | 104.6 (80) | 0.0 (110) | 0.0 | 3.1 | 8.8 |
| insurance.27.nb.(10-20) | 60 | $2^{14.0}$ | 1492.3 (18) | 5.0 (60) | 7.2 | 216.0 | 163.2 |
| insurance.27.nb.(20-40) | 10 | $2^{26.0}$ | – | 272.9 (10) | 741.9 | 4379.7 | 82.0 |
| poly.100.(0-10) | 95 | $2^{3.4}$ | 22.2 (62) | 0.0 (95) | 0.0 | 1.3 | 1.5 |
| poly.100.(10-20) | 5 | $2^{13.8}$ | – | 0.0 (5) | 0.0 | 95.0 | 3.0 |
| poly.100.nb.(0-10) | 83 | $2^{4.5}$ | 1.0 (48) | 0.0 (83) | 0.0 | 2.0 | 1.8 |
| poly.100.nb.(10-20) | 13 | $2^{15.2}$ | – | 0.0 (13) | 0.0 | 6.5 | 4.2 |
| poly.100.nb.(20-40) | 3 | $2^{28.7}$ | – | 1.0 (3) | 1.0 | 204.7 | 13.0 |
| poly.50.(0-10) | 81 | $2^{4.3}$ | 14.5 (55) | 0.0 (81) | 0.0 | 1.9 | 1.8 |
| poly.50.(10-20) | 16 | $2^{16.0}$ | – | 0.0 (16) | 0.0 | 11.9 | 4.5 |
| poly.50.(20-40) | 3 | $2^{25.3}$ | – | 0.0 (3) | 0.0 | 30.0 | 5.7 |
| rand.100.(0-10) | 31 | $2^{3.3}$ | 0.0 (20) | 0.0 (31) | 0.0 | 1.4 | 1.5 |
| rand.100.(10-20) | 10 | $2^{14.3}$ | – | 0.0 (10) | 0.0 | 8.5 | 4.2 |
| rand.100.(20-40) | 6 | $2^{24.0}$ | – | 1.2 (6) | 3.7 | 362.2 | 22.2 |
| rand.100.tw4.(0-10) | 49 | $2^{5.0}$ | 1.1 (22) | 0.0 (49) | 0.0 | 6.5 | 2.3 |
| rand.100.tw4.(10-20) | 26 | $2^{14.9}$ | – | 0.5 (26) | 13.8 | 477.5 | 6.0 |
| rand.100.tw4.(20-40) | 13 | $2^{26.1}$ | – | 8.8 (13) | 152.7 | 920.4 | 6.8 |
| rand.100.tw4.($>40$) | 2 | $2^{48.5}$ | – | 16.0 (2) | 44.0 | 557.5 | 10.0 |
| rand.100.tw8.(0-10) | 36 | $2^{3.9}$ | 90.7 (23) | 0.0 (36) | 0.0 | 6.5 | 2.1 |
| rand.100.tw8.(10-20) | 26 | $2^{15.5}$ | – | 6.8 (26) | 20.9 | 507.1 | 16.3 |
| rand.100.tw8.(20-40) | 19 | $2^{26.9}$ | – | 60.4 (19) | 451.6 | 1499.6 | 29.4 |
| rand.100.tw8.($>40$) | 2 | $2^{45.5}$ | – | 471.0 (2) | 2094.5 | 2635.5 | 39.5 |
| rand.30.(0-10) | 45 | $2^{4.5}$ | 196.4 (45) | 0.0 (45) | 0.0 | 2.3 | 2.4 |
| rand.30.(10-20) | 31 | $2^{15.2}$ | 649.1 (13) | 9.8 (31) | 105.1 | 862.0 | 40.2 |
| rand.30.(20-40) | 4 | $2^{21.8}$ | – | 908.0 (4) | 1569.0 | 6431.5 | 207.2 |
| rand.30.nb.(0-10) | 29 | $2^{3.9}$ | 9.9 (21) | 0.0 (29) | 0.0 | 2.7 | 1.8 |
| rand.30.nb.(10-20) | 16 | $2^{17.1}$ | 1338.0 (1) | 1.7 (16) | 2.3 | 336.8 | 15.5 |
| rand.30.nb.(20-40) | 11 | $2^{25.5}$ | – | 189.0 (11) | 638.7 | 3708.5 | 48.9 |
| rand.50.(0-10) | 35 | $2^{4.2}$ | 37.2 (21) | 0.0 (35) | 0.0 | 1.9 | 2.2 |
| rand.50.(10-20) | 20 | $2^{15.9}$ | – | 4.2 (20) | 9.8 | 490.3 | 21.6 |
| rand.50.(20-40) | 8 | $2^{25.1}$ | – | 71.2 (8) | 942.8 | 5389.1 | 103.6 |
| rand.50.nb.(0-10) | 83 | $2^{4.1}$ | 63.7 (61) | 0.0 (83) | 0.0 | 1.9 | 1.7 |
| rand.50.nb.(10-20) | 12 | $2^{15.5}$ | – | 0.0 (12) | 0.2 | 84.8 | 4.6 |
| rand.50.nb.(20-40) | 4 | $2^{26.5}$ | – | 0.0 (4) | 0.0 | 39.0 | 8.0 |
| rand.50.tw4.(0-10) | 47 | $2^{4.4}$ | 82.0 (27) | 0.0 (47) | 0.0 | 3.3 | 2.2 |
| rand.50.tw4.(10-20) | 27 | $2^{15.9}$ | – | 2.6 (27) | 44.5 | 642.1 | 5.5 |
| rand.50.tw4.(20-40) | 16 | $2^{27.9}$ | – | 39.1 (16) | 741.4 | 2861.4 | 8.2 |
| rand.50.tw8.(0-10) | 38 | $2^{4.9}$ | 3.3 (24) | 0.0 (38) | 0.0 | 4.0 | 2.5 |
| rand.50.tw8.(10-20) | 28 | $2^{15.7}$ | – | 2.7 (28) | 52.5 | 598.1 | 17.9 |
| rand.50.tw8.(20-40) | 20 | $2^{23.5}$ | – | 129.9 (20) | 274.4 | 890.0 | 33.9 |
| rand30iw4.(0-10) | 57 | $2^{5.0}$ | 241.6 (57) | 0.0 (57) | 0.0 | 7.3 | 2.5 |
| rand30iw4.(10-20) | 41 | $2^{15.0}$ | 1245.1 (26) | 0.3 (41) | 1.2 | 144.6 | 6.8 |
| rand30iw4.(20-40) | 1 | $2^{22.0}$ | – | 1.0 (1) | 2.0 | 118.0 | 10.0 |

Table 5: Average results of runs of the algorithms in many random generated networks where the new exact method failed to solve the problem (consequently SamIam also failed, as it has solved a subset of the cases that the new exact method could solve). #Q means number of queries. SS means search space size. Within parenthesis and near to time results are the counts of successful runs of the approximation algorithm.

| Net type | #Q | SS | Approx. time(sec) |
|---|---|---|---|
| insurance.27.nb.(20-40) | 20 | $2^{33.0}$ | – |
| poly.100.nb.(> 40) | 1 | $2^{50.0}$ | 0.0 (1) |
| rand.100.(10-20) | 5 | $2^{18.4}$ | – |
| rand.100.(20-40) | 17 | $2^{29.9}$ | 1222.7 (3) |
| rand.100.(> 40) | 28 | $2^{55.6}$ | – |
| rand.100.tw4.(20-40) | 6 | $2^{28.5}$ | 1434.2 (4) |
| rand.100.tw4.(> 40) | 4 | $2^{48.8}$ | 837.5 (2) |
| rand.100.tw8.(20-40) | 12 | $2^{32.0}$ | 1057.0 (4) |
| rand.100.tw8.(> 40) | 5 | $2^{47.0}$ | 85.0 (1) |
| rand.30.(10-20) | 1 | $2^{20.0}$ | 1137.0 (1) |
| rand.30.(20-40) | 19 | $2^{24.3}$ | 263.0 (3) |
| rand.30.nb.(20-40) | 11 | $2^{31.6}$ | 1009.3 (3) |
| rand.30.nb.(> 40) | 17 | $2^{48.1}$ | – |
| rand.50.(20-40) | 30 | $2^{33.8}$ | 1288.5 (2) |
| rand.50.(> 40) | 7 | $2^{42.4}$ | – |
| rand.50.nb.(20-40) | 1 | $2^{32.0}$ | 0.0 (1) |
| rand.50.tw4.(10-20) | 1 | $2^{20.0}$ | 11.0 (1) |
| rand.50.tw4.(20-40) | 9 | $2^{28.1}$ | 289.5 (4) |
| rand.50.tw8.(20-40) | 14 | $2^{28.2}$ | 943.4 (8) |
| rand30iw4.(20-40) | 1 | $2^{22.0}$ | 3537.0 (1) |

The main bottleneck of the algorithm is shared by many methods: the treewidth. While the constants and exponents that are hidden by the asymptotic notation in the analysis of the exact method are less aggressive than they look to be in a first moment, the complexity result of the multiplicative FPTAS might be seen at first as theoretical, because the number of hypercubes that are used to divide the vector space (given by Eq. (9)) is huge. Still it must be noted that the implementation of the FPTAS idea is not asymptotically slower than that of the exact algorithm, as we always keep only a subset of the full pareto set that is used at each level of the computation of the exact algorithm, but the division is so granulated that the number of discarded vectors (belonging to the same hypercube) is small in many cases, and the overhead to discard them might not be always computationally attractive. Still, Table 5 presents the cases that were not solved by the exact methods but were solved by the approximation method within one hour, which shows that the approximation algorithm can go beyond what the exact method can do in some cases.

In summary, this paper closes a few theoretical gaps with respect to the MAP problem in Bayesian networks and presents an efficient algorithm compared to currently available methods. It is also shown that a good approximation with theoretical guarantees is possible, but it is necessary to work on reducing the hidden constants and exponents of it. This is a point to be addressed in future work, as well as trying to devise an efficient pseudo-polynomial time algorithm (in fact, both ideas are strongly correlated). Another possibility is to study (theoretically and empirically) how to select vectors from the pareto sets in order to further reduce their sizes, which may produce very good approximation results in short time (but eventually without theoretical guarantees).

## Appendix A. Additional results used in the proofs of theorems

We present here the proofs that are not central to the arguments of the theorems of this paper. In fact they describe some simple mathematical relations that are exploited to ensure that the complexity reductions are performed in polynomial time.

**Lemma 14** *Let $k \geq 1$ be an integer and $f : \mathcal{N} \to \mathcal{Q}$ and $g : \mathcal{N} \to \mathcal{N}$ two functions such that $f(k) \geq 1 + \log_2 g(k)$ and $g(k) \geq 1$. Then $2^{-f(k)} < 2^{\frac{1}{g(k)}} - 1$.*

**Proof** First note that $2^{-i \cdot f(k)} \leq (2^{\log_2 2g(k)})^{-i} = \frac{1}{(2g(k))^i}$ for any $0 \leq i \leq g(k)$, and that $\binom{g(k)}{i} \leq g(k)^i$. Now,

$$(2^{-f(k)} + 1)^{g(k)} = \sum_{i=0}^{g(k)} \binom{g(k)}{i} \cdot 2^{-i \cdot f(k)} \leq \sum_{i=0}^{g(k)} g(k)^i \cdot 2^{-i \cdot f(k)}$$

$$\leq \sum_{i=0}^{g(k)} g(k)^i \frac{1}{(2g(k))^i} = \sum_{i=0}^{g(k)} \frac{1}{2^i} < \sum_{i=0}^{\infty} \frac{1}{2^i} = 2.$$

Finally,
$$(2^{-f(k)} + 1)^{g(k)} < 2 \Rightarrow (2^{-f(k)} + 1) < 2^{\frac{1}{g(k)}} \Rightarrow 2^{-f(k)} < 2^{\frac{1}{g(k)}} - 1. \square$$

**Lemma 15** *Let $v \geq 2^{-b}$ be a rational number and $k \geq 1$ an integer. Then $v + 2^{-(k+b+1)} < v \cdot 2^{\frac{1}{2^k}}$ and $v - 2^{-(k+b+\frac{3}{2})} > v \cdot 2^{-\frac{1}{2^k}}$.*

**Proof** The first inequality follows from Lemma 14 with $f(k) = k+1$ and $g(k) = 2^k$:
$$2^{-(k+1)} < 2^{\frac{1}{2^k}} - 1 \Rightarrow 2^{-(k+1)} \cdot 2^{-b} < v \cdot (2^{\frac{1}{2^k}} - 1) \Rightarrow v + 2^{-(k+b+1)} < v \cdot 2^{\frac{1}{2^k}}$$

The other result is analogous using $f(k) = k - \frac{1}{2^k} + \frac{3}{2}$ and $g(k) = 2^k$:
$$2^{-(k-\frac{1}{2^k}+\frac{3}{2})} < 2^{\frac{1}{2^k}} - 1 \Rightarrow 2^{-(k+\frac{3}{2})} < \frac{2^{\frac{1}{2^k}} - 1}{2^{\frac{1}{2^k}}} \Rightarrow 2^{-(k+\frac{3}{2})} < 1 - 2^{-\frac{1}{2^k}} \Rightarrow$$
$$2^{-(k+\frac{3}{2})-b} < v \cdot (1 - 2^{-\frac{1}{2^k}}) \Rightarrow v - 2^{-(k+b+\frac{3}{2})} > v \cdot 2^{-\frac{1}{2^k}}. \square$$

**Corollary 16** $2^{-v} + 2^{-(k+3)} < 2^{-v+\frac{1}{2^k}}$ *and* $2^{-v} - 2^{-(k+4)} > 2^{-v-\frac{1}{2^k}}$ *for every $v \leq 2$ and integer $k \geq 1$.*

**Proof** It follows from Lemma 15 with $b = 2$. $\square$

**Lemma 17** *Given a rational $0 \leq x \leq \frac{1}{2}$, we have that $\log_2(1 + 2^{2x}) - x^4 - x - 1 \geq 0$.*

**Proof** We apply a Taylor expansion around zero for the expression as follows:
$$\log_2(1 + 2^{2x}) - x^4 - x - 1 \in \left( \frac{\log 2}{2} x^2 - \frac{12 + (\log 2)^3}{12} x^4 + O(x^5) \right).$$

And then some manipulations are enough to show the desired result:
$$\frac{\log 2}{2} - \frac{12 + (\log 2)^3}{48} \geq 0 \Longrightarrow \frac{\log 2}{2} x^2 - \frac{12 + (\log 2)^3}{48} x^2 \geq 0 \Longrightarrow$$
$$\frac{\log 2}{2} x^2 - \frac{1}{4} \frac{12 + (\log 2)^3}{12} x^2 \geq 0 \Longrightarrow \frac{\log 2}{2} x^2 - \frac{12 + (\log 2)^3}{12} x^4 \geq 0 \Longrightarrow \log_2(1 + 2^{2x}) - x^4 - x - 1 \geq 0. \square$$